\documentclass[runningheads]{llncs}

\usepackage[mobile]{eccv}

\usepackage[dvipsnames]{xcolor}

\usepackage{booktabs, makecell, tabularx}
\usepackage{multirow}

\usepackage{multirow}
\usepackage{arydshln}

\usepackage{tikz}
\usetikzlibrary{shapes.arrows}

\newcommand{\refsec}[1]{Section~\ref{sec:#1}}

\newcommand{\refeq}[1]{Equation~\ref{eq:#1}}

\newcommand{\lblsec}[1]{\label{sec:#1}}
\newcommand{\lbleq}[1]{\label{eq:#1}}

\newcommand{\myparagraph}[1]{\vspace{-2pt} \smallskip \noindent \textbf{#1}}

\newcommand{\methodname}{\texttt{CycleGAN-Turbo}\xspace}
\newcommand{\pairedname}{\texttt{pix2pix-Turbo}\xspace}

\newcommand{\inputDomain}{$\mathcal{X} \subset \mathbb{R}^{H \times W \times 3}$~}
\newcommand{\inputImage}{$x$\xspace}
\newcommand{\inputNoise}{$z$\xspace}

\newcommand{\outputDomain}{$\mathcal{Y} \subset \mathbb{R}^{H \times W \times 3}$~}
\newcommand{\unpairedDataset}{$X=\{x \in \mathcal{X} \}, ~Y= \{y \in \mathcal{Y} \}$~}

\newcommand{\G}{G}

\newcommand{\captionY}{c_Y}
\newcommand{\captionX}{c_X}
\newcommand{\strength}{\gamma}
\newcommand{\Lrec}{\mathcal{L}_{\text{rec}}}

\newcommand{\LCLIP}{\mathcal{L}_{\text{CLIP}}}
\newcommand{\DiscX}{\mathcal{D}_X}
\newcommand{\DiscY}{\mathcal{D}_Y}

\definecolor{myblue}{rgb}{0.239,0.553,0.565}

\definecolor{tabred}{RGB}{255,77,0}
\definecolor{taborange}{RGB}{255,154,0}
\definecolor{tabgreen}{RGB}{8, 143, 143}

\usepackage{eccvabbrv}

\usepackage[accsupp]{axessibility}  %

\usepackage[pagebackref,breaklinks,colorlinks,citecolor=eccvblue]{hyperref}

\author{Gaurav Parmar\inst{1} \quad Taesung Park\inst{2} \quad Srinivasa Narasimhan\inst{1} \quad Jun-Yan Zhu\inst{1}}

\institute{Carnegie Mellon University\inst{1}\; \;
Adobe Research\inst{2}\; \;  }

\usepackage{orcidlink}

\begin{document}

\title{One-Step Image Translation with \\Text-to-Image Models
\vspace{-0.1in}}
\titlerunning{One-Step Image Translation with Text-to-Image Models}

\maketitle
\vspace{-0.1in}
\begin{abstract}

In this work, we address two limitations of existing conditional diffusion models: their slow inference speed due to the iterative denoising process and their reliance on paired data for model fine-tuning. To tackle these issues,  
we introduce a general method for adapting a single-step diffusion model to new tasks and domains through adversarial learning objectives. Specifically, we consolidate various modules of the vanilla latent diffusion model into a single end-to-end generator network with small trainable weights, enhancing its ability to preserve the input image structure while reducing overfitting. 
We demonstrate that, for unpaired settings, our model \methodname outperforms existing GAN-based and diffusion-based methods for various scene translation tasks, such as day-to-night conversion and adding/removing weather effects like fog, snow, and rain. We extend our method to paired settings, where our model \pairedname is on par with recent works like ControlNet for Sketch2Photo and Edge2Image, but with a single-step inference. This work suggests that single-step diffusion models can serve as strong backbones for a range of GAN learning objectives.
 Our code and models are available at \url{https://github.com/GaParmar/img2img-turbo}. 
\end{abstract}

\begin{figure}
    \centering 
    \includegraphics[width=\linewidth]{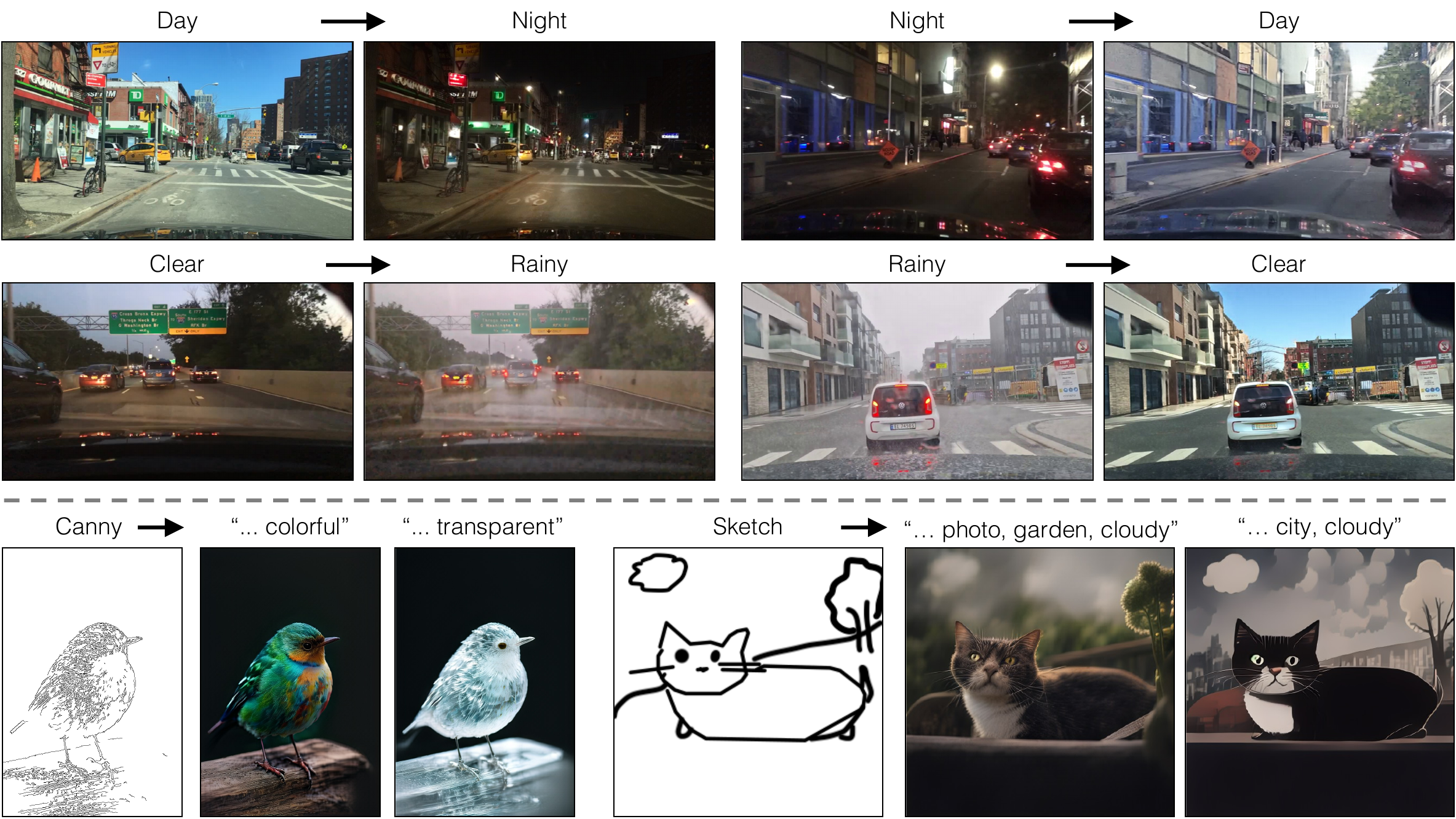}
        \vspace{-10pt}
    \captionof{figure}{
We present a general method for adapting a single-step diffusion model, such as  SD-Turbo~\cite{sauer2023adversarial}, to new tasks and domains through adversarial learning. This enables us to leverage the internal knowledge of pre-trained diffusion models while achieving efficient inference (e.g., 0.3 seconds for 512x512 image). Our single-step image-to-image translation models, called \methodname and \pairedname, can synthesize realistic outputs for unpaired  (top) and paired settings (bottom), respectively, on various tasks.
    } 
    \label{fig:teaser}
    \vspace{-15pt}
\end{figure}

\vspace{-0.3in}
\section{Introduction} \label{sec:intro}

Conditional diffusion models~\cite{zhang2023adding, brooks2022instructpix2pix, saharia2022palette,mou2023t2i} have empowered users to generate images based on both spatial conditioning and text prompts, enabling various image synthesis applications that demand precise user controls over scene layout, user sketches, and human poses. Despite their huge success, these models face two primary challenges. First, the iterative nature of diffusion models makes inference slow, limiting real-time applications,  such as interactive Sketch2Photo. Second, model training often requires curating large-scale paired datasets, posing significant costs for many applications, while being infeasible for others~\cite{zhu2017unpaired}.

In this work, we introduce a one-step image-to-image translation method applicable to both paired and unpaired settings. Our method achieves visually appealing results comparable to existing conditional diffusion models, while reducing the number of inference steps to 1. More importantly, our method can be trained without image pairs.  Our key idea is to efficiently adapt a pre-trained text-conditional one-step diffusion model, such as SD-Turbo~\cite{sauer2023adversarial},  to new domains and tasks via adversarial learning objectives. %

Unfortunately, directly applying standard diffusion adapters like ControlNet~\cite{zhang2023adding} 
to the one-step setting proved less effective in our experiments. Unlike traditional diffusion models, we observe that the noise map directly influences the output structure in the one-step model. Consequently, feeding both noise maps and input conditioning through additional adapter branches results in conflicting information for the network. Especially for unpaired cases, this strategy leads to the original network being disregarded by the end of training. Moreover, many visual details in the input image are lost during image-to-image translation, due to imperfect reconstruction by the multi-stage pipeline (Encoder-UNet-Decoder) of the SD-Turbo model. This loss of detail is particularly noticeable and crucial when the input is a real image, such as in day-to-night translation.

To tackle these challenges, we propose a new generator architecture that leverages SD-Turbo weights while preserving the input image structure. First, we feed the conditioning information directly to the noise encoder branch of the UNet. This enables the network to adapt to new controls directly, avoiding conflicts between the noise map and the input control. Second, we consolidate the three separate modules, Encoder, UNet, and Decoder, into a single end-to-end trainable architecture. For this, we employ LoRA~\cite{hu2021lora} to adapt the original network to new controls and domains,  reducing overfitting and fine-tuning time. Finally, to preserve the high-frequency details of the input, we incorporate skip connections between the encoder and decoder via zero-conv~\cite{zhang2023adding}. Our architecture is versatile, serving as a plug-and-play model for conditional GAN learning objectives such as CycleGAN and pix2pix~\cite{zhu2017unpaired,isola2017image}. To our knowledge, our work is the first to achieve one-step image translation with a text-to-image model. 

We primarily focus on the harder unpaired translation tasks, such as converting from day to night and vice versa and adding/removing weather effects to/from images. We show that our model  \methodname significantly outperforms both existing GANs-based and diffusion-based methods in terms of distribution matching and input structure preservation, while achieving greater efficiency than diffusion-based methods. We include an extensive ablation study regarding each design choice of our method.  

To demonstrate the versatility of our architecture, we also perform experiments for paired settings, such as Edge2Image or Sketch2Photo. Our model called \pairedname achieves visually comparable results with recent conditional diffusion models, while reducing the number of inference steps to 1. We can generate diverse outputs by interpolating between noise maps used in pre-trained model and our model's encoder outputs. In summary, our work suggests that one-step pre-trained text-to-image models can serve as a strong and versatile backbone for many downstream image synthesis tasks.

\vspace{-0.1in}
\section{Related Work} \label{sec:related_works}
\vspace{-0.08in}
\noindent \textbf{Image-to-Image translation.}
Recent advances in generative models have enabled many image-to-image translation applications. Paired image translation methods~\cite{isola2017image,sangkloy2016scribbler,park2019semantic,wang2018pix2pixHD,zhao2021large,zhu2020sean} map an image from a source domain to a target domain, using a combination of reconstruction~\cite{johnson2016perceptual,zhang2018unreasonable} and adversarial losses~\cite{goodfellow2014generative}. More recently, various conditional diffusion models have emerged, integrating text and spatial conditions for image translation tasks~\cite{avrahami2023spatext,wang2022pretraining,zhang2023adding,li2023gligen,mou2023t2i,brooks2022instructpix2pix,saharia2022palette}. These methods often build upon pre-trained text-to-image models. For instance, works like GLIGEN~\cite{li2023gligen}, T2I-Adapter~\cite{mou2023t2i}, and ControlNet~\cite{zhang2023adding} introduce effective fine-tuning techniques using adapters such as gated transformer layers or zero-convolution layers. However, the model training still requires a large number of training pairs. In contrast, our approach can leverage large-scale diffusion models without image pairs, with significantly faster inference speed. 

In many cases where paired input and output images are unavailable, several techniques have been proposed, including cycle consistency~\cite{zhu2017unpaired, yi2017dualgan, kim2017learning}, shared intermediate latent space~\cite{liu2017unsupervised, huang2018munit,lee2018diverse}, content preservation loss~\cite{shrivastava2017learning,taigman2017unsupervised}, and contrastive learning~\cite{park2020contrastive,han2021dual}. Recent works ~\cite{cyclediffusion, su2022dual, sasaki2021unit_ddpm} have also explored diffusion models for unpaired translation tasks. 
However, these GAN-based or diffusion-based methods typically require training from scratch on new domains. %
Instead, we introduce the first unpaired learning method leveraging pre-trained diffusion models, demonstrating better results than existing methods.

\myparagraph{Text-to-Image models.}
Large-scale text-conditioned models~\cite{balaji2022eDiff-I, ramesh2022hierarchical, nichol2022glide, saharia2022photorealistic, gafni2022make, kang2023gigagan} have significantly improved image quality and diversity through training on internet-scale datasets~\cite{schuhmann2022laion, kakaobrain2022coyo-700m}.
Several works~\cite{meng2022sdedit, hertz2022prompt, plug_and_play_Tumanyan, parmar2023zero, mokady2023null} have proposed zero-shot methods for editing real images with pre-trained text-to-image models. For example, SDEdit~\cite{meng2022sdedit} edits real images by adding noise to the input image and subsequently denoises with a pre-trained model according to the text prompt. Prompt-to-Prompt works further manipulate or preserve features in cross-attention and self-attention layers during the image editing process~\cite{hertz2022prompt, plug_and_play_Tumanyan, parmar2023zero,chefer2023attend,ge2023expressive,patashnik2023localizing,cao2023masactrl}. %
Others fine-tune the networks or text embeddings for the input image before image editing~\cite{kawar2023imagic,mokady2023null} or employ more precise inversion methods~\cite{song2020denoising,wallace2023edict}. 
Despite their impressive results, they frequently encounter difficulties in complex scenes with many objects. Our work can be viewed as augmenting these methods with paired or unpaired data from new domains/tasks.

\myparagraph{One-step generative models.}
To expedite diffusion model inference, recent works focus on reducing the number of sampling steps using fast ODE solvers~\cite{lu2022dpm,karras2022elucidating}, or distilling slow multistep teacher models into fast few-step student models~\cite{meng2023distillation,salimans2022progressive}. Regressing directly from noise to images often produces blurry results~\cite{luhman2021knowledge,zheng2023fast}. For this, various distillation methods use consistency model training~\cite{luo2023latent,song2023consistency}, adversarial learning~\cite{sauer2023adversarial,xu2023ufogen}, variational score distillation~\cite{yin2023one,wang2024prolificdreamer}, Rectified Flow~\cite{liu2022flow,liu2023instaflow}, and their combinations~\cite{sauer2023adversarial}. Other methods directly use GANs for text-to-image synthesis~\cite{kang2023gigagan,sauer2023stylegan}.  Different from these works that focus on one-step text-to-image synthesis, we present the first one-step conditional model that use both text and conditioning images. Our method beats the baseline that directly uses the original ControlNet with one-step distilled models.

\begin{figure}[t!]
    \centering
    \includegraphics[width=\linewidth]{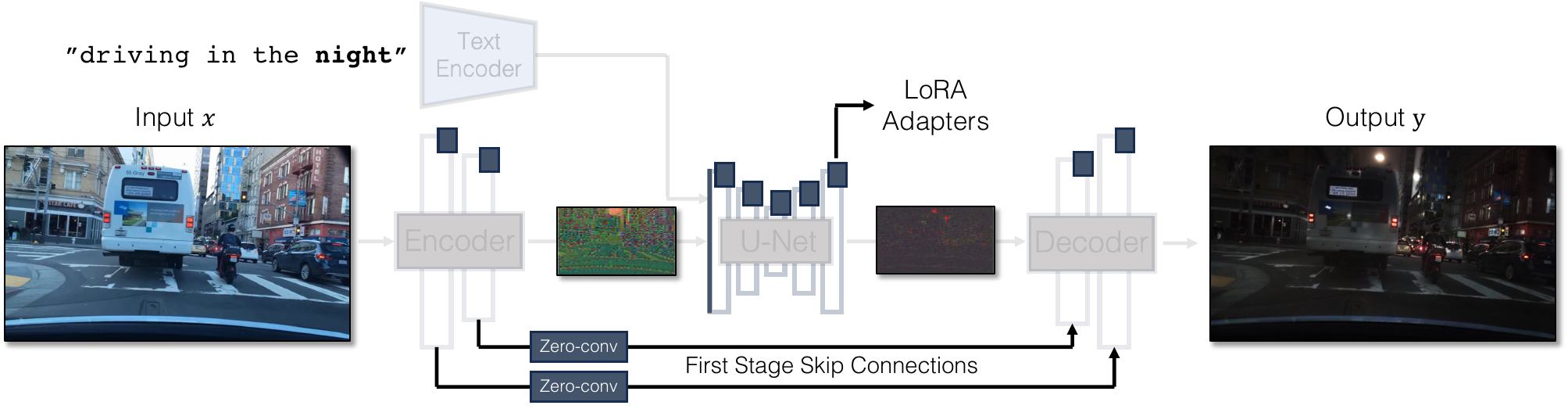}
        \vspace{-10pt}
    \caption{{\bf Our generator architecture.} We tightly integrate three separate modules in the original latent diffusion models into a single end-to-end network with small trainable weights. This architecture allows us to translate the input image $x$ to the output $y$, while retaining the input scene structure.  We use LoRA adapters~\cite{hu2021lora} in each module, introduce skip connections and Zero-Convs~\cite{zhang2023adding} between input and output, and retrain the first layer of the U-Net.  Blue boxes indicate trainable layers. Semi-transparent layers are frozen. The same generator can be used for various GAN objectives. 
    }
    \label{fig:pipeline}
            \vspace{-15pt}
\end{figure}
\vspace{-0.15in}
\section{Method} \label{sec:method}
\vspace{-0.1in}
We start with a one-step pre-trained text-to-image model capable of generating realistic images.
However, our goal is to \textit{translate} an input real image from a source domain to a target domain, such as converting a day driving image to night.
In Section~\ref{sec:adding_structure}, we explore different conditioning methods for adding structure to our model and the corresponding challenges. 
Next, in Section~\ref{sec:preserving_details}, we investigate the common issue of detail loss (e.g., text, hands, street signs) that plagues latent-space models~\cite{rombach2022high} and propose a solution to address it. 
We then discuss our unpaired image translation method in \refsec{training}, with further extensions to paired settings and stochastic generation (Section~\ref{sec:extension}).
\begin{figure}[t!]
    \centering
    \includegraphics[width=1.0\linewidth]{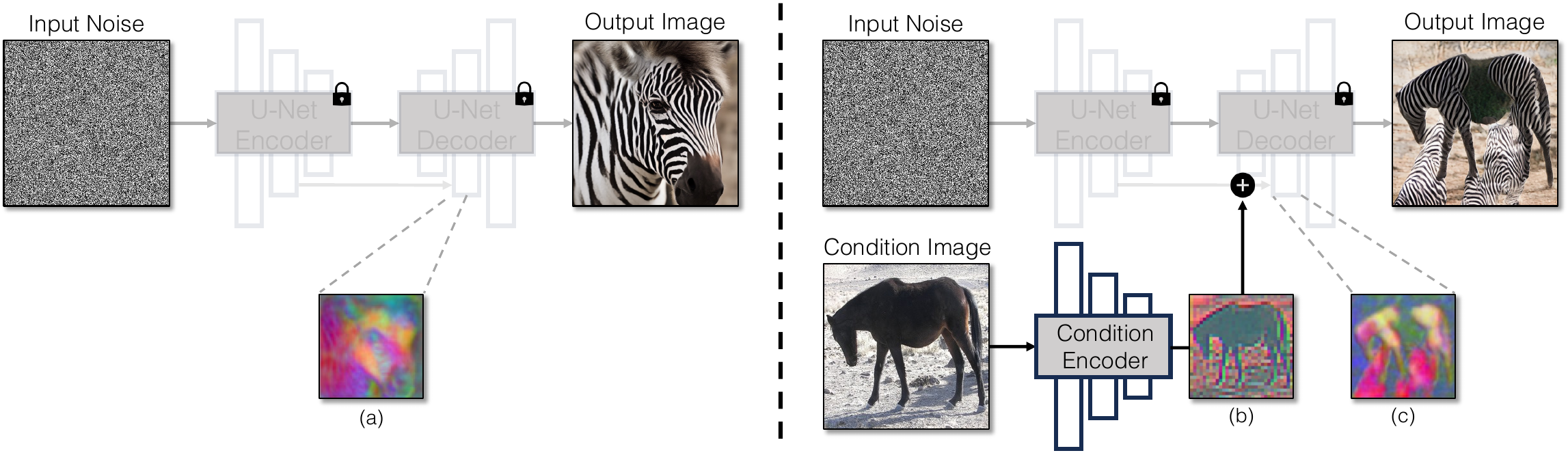}
     \vspace{-10pt}
    \caption{
    \textit{(Left)} The one-step model learns to map the input noise to the output image. Note that the features of SD2.1-Turbo forms a coherent layout (a) from the noise map. \textit{(Right)} Unfortunately, adding condition encoder branches~\cite{zhang2023adding,mou2023t2i} causes conflicts, since features (b) from the new branch represent a different layout compared to the original feature (a). This conflict deteriorates the downstream feature (c) in the SD-Turbo Decoder, affecting the output quality.  The feature maps are visualized with PCA. 
    }
     \vspace{-15pt}
    \label{fig:conditioning_conflict}
\end{figure}

\subsection{Adding Conditioning Input}\label{sec:adding_structure}
To convert a text-to-image model into an image translation model, we first need to find an effective way to incorporate the input image \inputImage into the model.

\noindent \textbf{Conflicts between noise and conditional input.}
One common strategy for incorporating conditional input into Diffusion models is introducing extra adapter branches~\cite{zhang2023adding,mou2023t2i}, as shown in Figure~\ref{fig:conditioning_conflict}. Concretely, we initialize a second encoder, labeled as the Condition Encoder, either with the weights of the Stable Diffusion Encoder~\cite{zhang2023adding} or using a lightweight network with randomly initialized weights~\cite{mou2023t2i}. This Control Encoder takes the input image \inputImage, and outputs feature maps at multiple resolutions to the pre-trained Stable Diffusion model through residual connections. 
This method has yielded remarkable outcomes for controlling diffusion models. 
Nonetheless, as illustrated in Figure~\ref{fig:conditioning_conflict}, using two encoders (U-Net Encoder and Condition Encoder) to process a noise map and an input image presents challenges in the context of one-step models. Unlike multi-step diffusion models, the noise map in the one-step model directly controls the layout and pose of generated images, often contradicting the structure of the input image. 
Hence, the decoder receives two sets of residual features, each representing distinct structures, making the training process more challenging.

\myparagraph{Direct conditioning input.}
Figure~\ref{fig:conditioning_conflict} also illustrates that the structure of the generated image by the pre-trained model is significantly influenced by the noise map \inputNoise. 
Based on this insight, we propose that the conditioning input should be fed to the network directly. Figure~\ref{fig:abl_all} and Table~\ref{tab:ablation_study} additionally show that using direct conditioning achieves better results than using an additional encoder. 
To allow the backbone model to adapt to new conditioning, we add several LoRA weights~\cite{hu2021lora} to various layers in the U-Net (see Figure~\ref{fig:pipeline}).

\subsection{Preserving Input Details}\label{sec:preserving_details}
A key challenge that prevents the use of latent diffusion models (LDM ~\cite{rombach2022high}) in multi-object and complex scenes is the lack of detail preservation.

\begin{figure}[t!]
    \centering
    \includegraphics[width=0.95\linewidth]{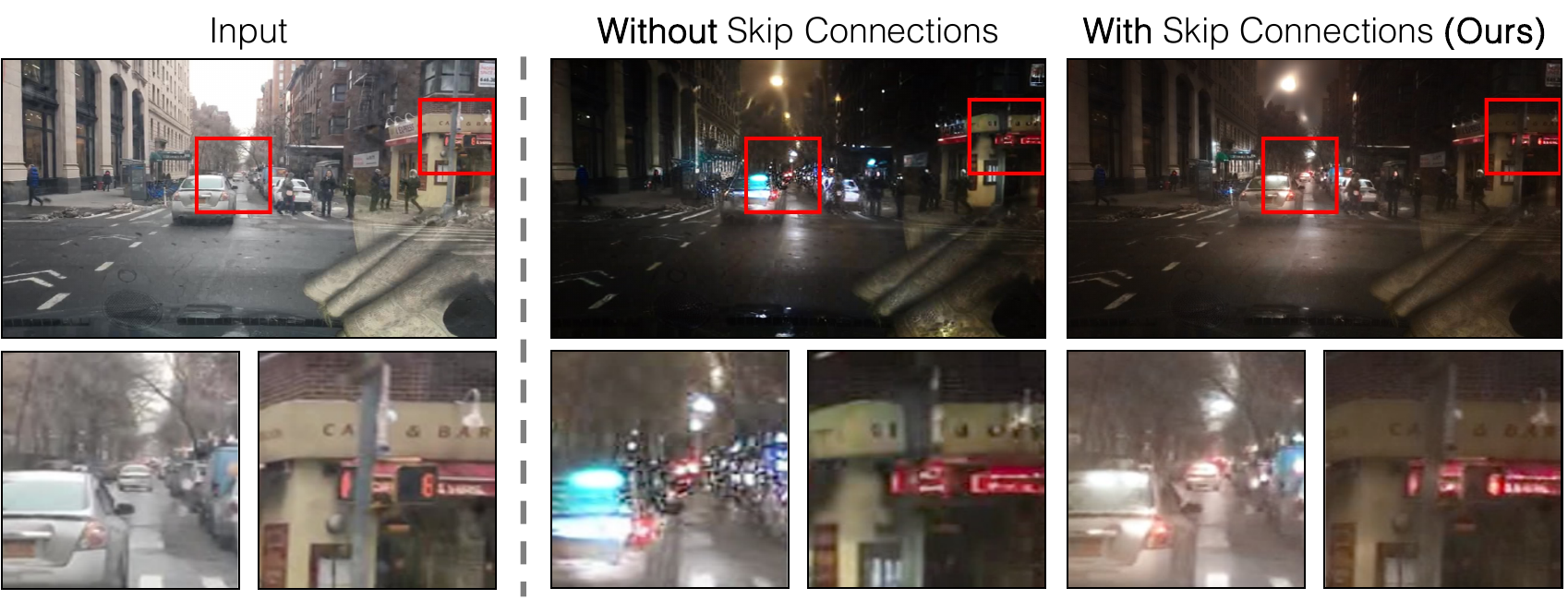}
     \vspace{-10pt}
    \caption{{\bf Skip Connections help retain details.} We visualize the outputs of our day-to-night models trained with and without skip connections. It is clearly seen that adding skip connections preserves the details of the input daytime image. The zoomed in crops of the night images are gamma-adjusted by 1.5 for easier visualization. }
    \label{fig:details_lost}
     \vspace{-15pt}
\end{figure}

\myparagraph{Why details are lost.}
The image encoder of Latent Diffusion Models (LDMs) compresses input images spatially by a factor of 8 while increasing the channel count from 3 to 4. This design speeds up the training and inference of diffusion models. However, it may not be ideal for image translation tasks, which require preserving fine details of the input image.
We illustrate this issue in Figure~\ref{fig:details_lost}, where we take an input daytime driving image (left) and translate it to a corresponding nighttime driving with an architecture that does not use skip connections (middle). Observe that fine-grained details, such as text, street signs, and cars in the distance, are not preserved. In contrast, employing an architecture that incorporates skip connections (right) results in a translated image that significantly better retains these intricate details. 

\myparagraph{Connecting first stage encoder and decoder.}
To capture fine-grained visual details of the input image, 
we add skip connections between the Encoder and Decoder networks (see Figure~\ref{fig:pipeline}). 
Specifically, we extract four intermediate activations following each downsampling block within the encoder, process them via a 1$\times$1 zero-convolution layer~\cite{zhang2023adding}, and then feed them into the corresponding upsampling block in the decoder. This method ensures the retention of intricate details throughout the image translation process.

\subsection{Unpaired Training}\label{sec:training}
We use Stable Diffusion Turbo (v2.1) with one-step inference as the base network for all of our experiments. Here we show that our generator can be used in a modified CycleGAN formulation~\cite{zhu2017unpaired} for unpaired translation. Concretely, we aim to
convert images from a source domain \inputDomain to some desired target domain \outputDomain, given an \textit{unpaired} dataset \unpairedDataset. 

Our method includes two translation functions $G(x, \captionY)$: $X \rightarrow Y$ and $G(y,\captionX)$: $Y \rightarrow X$. 
Both translations use the same network $\G$ as described in \refsec{adding_structure} and \refsec{preserving_details}, but different captions $\captionX$ and $\captionY$ that correspond to the task. For example, in the day $\rightarrow$ night translation task, $\captionX$ is \texttt{Driving in the day}, and $\captionY$ is \texttt{Driving in the night}.  As depicted in Figure~\ref{fig:pipeline}, we keep most layers frozen and only train the first convolutional layer and the added LoRA adapters.

\myparagraph{Cycle consistency with perceptual loss.}
The cycle consistency loss $\mathcal{L}_\text{cycle}$ enforces that for each source image \inputImage, the two translation functions should bring it back to itself. We denote $\Lrec$ a combination of L1 difference and LPIPS~\cite{zhang2018unreasonable}. Please refer to Appendix~\ref{sup_sec:training_details} for the weighting. 
\begin{equation}\lbleq{cycle_loss}
    \begin{aligned}
        \mathcal{L}_{\text{cycle}} = \mathbb{E}_{x} \left[   \Lrec(\G(\G(x, \captionY), \captionX), x) \right] \\
         + ~ \mathbb{E}_{y} \left[\Lrec(\G(\G(y, \captionX), \captionY), y)\right]
    \end{aligned}
\end{equation}

\myparagraph{Adversarial loss.} 
We use an adversarial loss \cite{goodfellow2014generative} for both domains to encourage the translated outputs to match the corresponding target domains. 
We use two adversarial discriminators, $\DiscX$ and $\DiscY$, that aim to classify real images from the translated images for the corresponding domains. Both discriminators use the CLIP model as a backbone, following the recommendations of Vision-Aided 
GAN~\cite{kumari2021ensembling}. The adversarial loss can be defined as: 
\begin{equation}
\lbleq{gan_loss}
\begin{aligned}
\mathcal{L}_\text{GAN} = \mathbb{E}_{y} \left[ \log \DiscY(y) \right] + \mathbb{E}_{x} \left[ \log (1 - \DiscY(G(x, \captionY)))\right] \\
+  \mathbb{E}_{x} \left[ \log \DiscX(x) \right] + \mathbb{E}_{y} \left[ \log (1 - \DiscX(G(y, \captionX)))\right]
\end{aligned}
\end{equation}

\myparagraph{Full objective.}
The complete training objective comprises of three different losses: cycle consistency loss $\mathcal{L}_\text{cycle}$, adversarial loss $\mathcal{L}_\text{GAN}$ and identity regularization loss $\mathcal{L}_\text{idt} =   \mathbb{E}_{y} \left[  \Lrec(\G(y, \captionY), y) \right] + \mathbb{E}_{x} \left[  \Lrec(\G(x, \captionX), x) \right]$. The loss is weighted by $\lambda_\text{idt}$ and $\lambda_\text{gan}$, as follows: 
\begin{equation}\lbleq{full_objective}\begin{aligned}
\arg \min_\G \mathcal{L}_{\text{cycle}} + \lambda_\text{idt}\mathcal{L}_{\text{idt}} + \lambda_\text{GAN} \mathcal{L}_\text{GAN}.
\end{aligned}\end{equation} 

\vspace{-15pt}
\subsection{Extensions}
\label{sec:extension}
While our primary focus is on unpaired learning, we also demonstrate two extensions to learn other types of GAN objectives, such as learning from paired data and generating stochastic outputs.

\myparagraph{Paired training.} \label{sec:paired_training}
We adapt our translation network $G$  to paired settings, such as converting edges or sketches to images. We refer to the paired version of our method as \pairedname.
In the paired setting, we aim to learn a single translation function  $G(x, c)$: $X \rightarrow Y$, where X is the source domain (e.g., input sketch), Y is the target domain (e.g., output image), and $c$ is the input caption. 
For paired training objective, we use (1) reconstruction loss as a combination of perceptual loss and pixel-space reconstruction loss,  (2) GAN loss, similar to the loss in \refeq{gan_loss}, but only for the target domain, and (3) CLIP text-image alignment loss $\LCLIP$~\cite{radford2021learning}. %
Please find more details in Appendix~\ref{sup_sec:training_details}. %

\myparagraph{Generating diverse outputs}
Generating stochastic outputs is important in many image translation tasks, e.g., sketch-to-image generation. However, enabling a one-step model to generate diverse outputs is challenging as it needs to make use of additional input noise, which often gets ignored~\cite{zhu2017toward,huang2018munit}. 
We propose generating diverse outputs by interpolating the features and model weights toward the pretrained model, which already produces diverse outputs. 
Concretely, given an interpolation coefficient $\strength$, we make the following three changes. First, we combine the Gaussian noise and the encoder output.  Our generator $G(x, z, r)$ now takes three inputs: the input image $x$, a noise map $z$, and the coefficient $\strength$. %
The updated function $G(x, z, \strength)$ first combines the noise $z$ and the encoder output: %
$\strength \; G_{\text{enc}}(x) + (1-\strength) \; z$. %
We then feed the combined signal to the U-Net. 

Second, we also scale the LoRA adapter weights and outputs of the skip connections according to $\theta = \theta_0 + \strength \cdot \Delta \theta$, where  $\theta_0$ and $ \Delta \theta$ denote the original weights and newly added weights, respectively. 

Finally, we scale the reconstruction loss according to the coefficient $\strength$.
\begin{equation}\lbleq{finetune_rec}
     \mathcal{L}_{\text{diverse}} = \mathbb{E}_{x, y, z, \strength} \left[ \strength \mathcal{L}_\text{rec}(G(x, z, \strength), y) \right].
\end{equation} 

Notably, $\strength=0$ corresponds to the default stochastic behavior of the pretrained model, in which case the reconstruction loss is not enforced. $\strength=1$ corresponds to the deterministic translation described in Sections~\ref{sec:training} and \ref{sec:paired_training}. 
We finetune our image translation models with varying interpolation coefficients. %
Figure~\ref{fig:diversity_paired} shows that such a finetuning enables our model to generate diverse outputs by sampling different noises during inference time. %

\begin{figure}[t!]
    \centering
    \includegraphics[width=\linewidth]{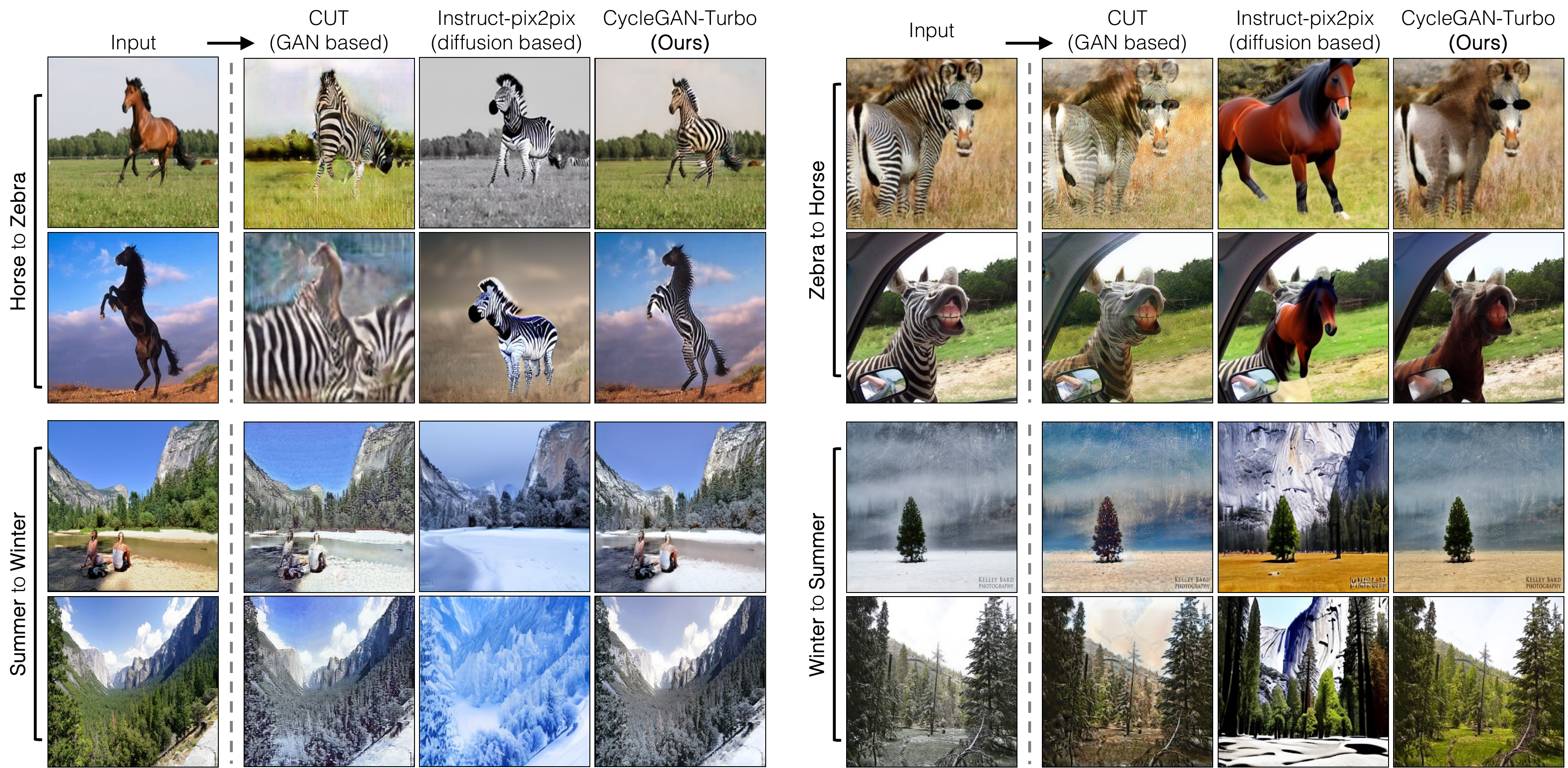}
       \vspace{-10pt}
    \caption{{\bf Comparison to baselines on 256 $\times$ 256 datasets.} 
    We compare our unpaired method to CUT~\cite{park2020contrastive} and Instruct-pix2pix~\cite{brooks2022instructpix2pix}, the best-performing GAN-based and diffusion methods, respectively.
    CUT outputs images that often contain severe image artifacts. Whereas, Instruct-pix2pix fails to preserve the input image structure. 
    }
    \label{fig:cmp_unpaired_small_sd}
    \vspace{-15pt}
\end{figure}

\begin{figure}[t!]
    \centering
    \includegraphics[width=\linewidth]{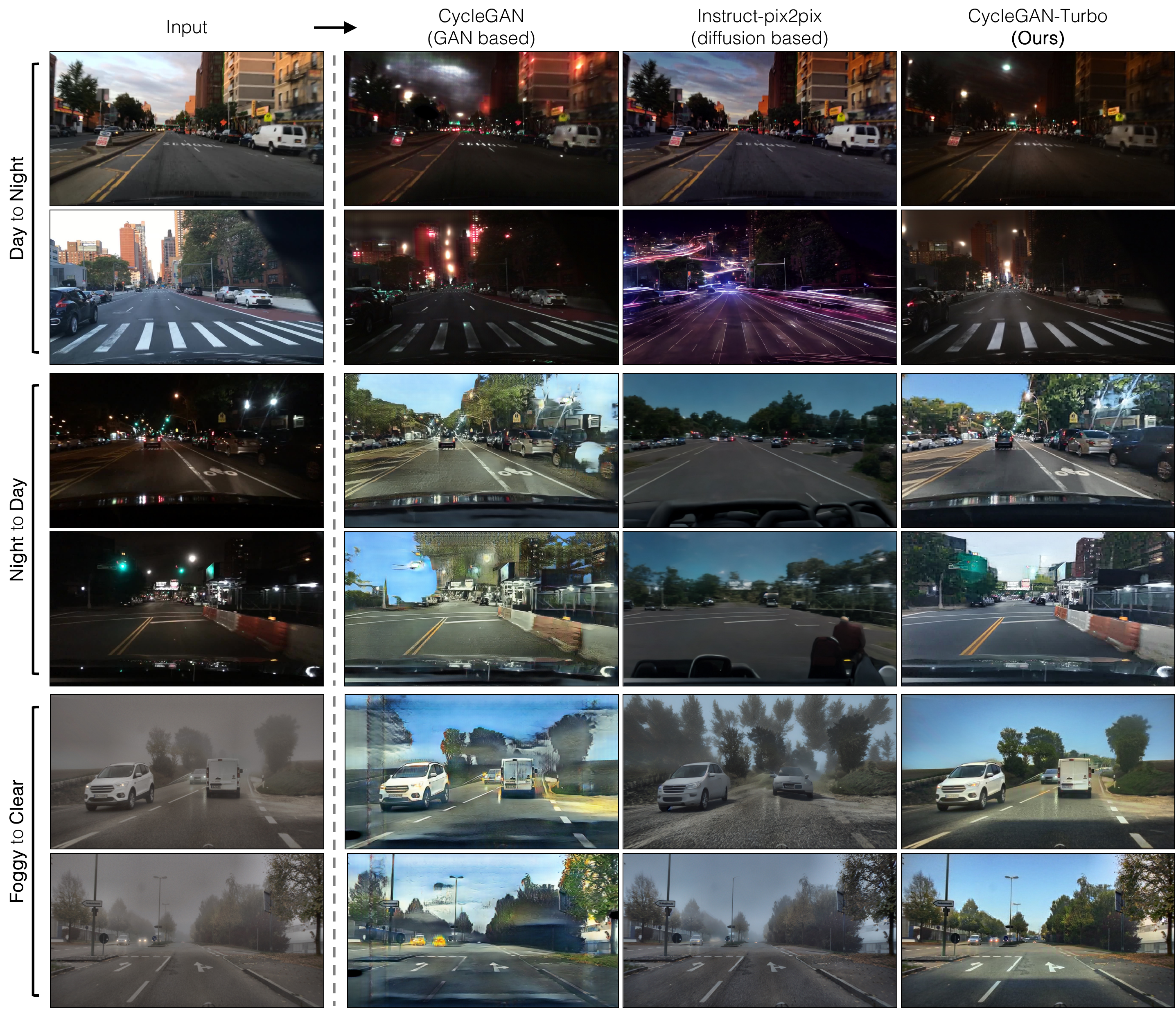}
       \vspace{-10pt}
    \caption{{\bf Comparison to baselines on driving datasets (512 $\times$ 512).}
    We compare our unpaired translation method to CycleGAN~\cite{zhu2017unpaired} and Instruct-pix2pix~\cite{brooks2022instructpix2pix}, the best performing GAN-based and diffusion methods for this dataset. CycleGAN does not use existing text-to-image models and, as a result, generates artifacts in the outputs, e.g., the sky regions in the day-to-night translation. In contrast, Instruct-pix2pix uses a large text-to-image model but does not use the unpaired dataset. So, the Instruct-pix2pix outputs look unnatural and vastly different than the images in our datasets.  
    }
    \label{fig:cmp_unpaired_driving_ds}
       \vspace{-15pt}
\end{figure}

\vspace{-10pt}
\section{Experiments} \label{sec:experiments}
We conduct extensive experiments on several image translation tasks, organized into three main categories. 
First, we compare our method to several prior GAN-based and diffusion model image translation methods, demonstrating better quantitative and qualitative results. 
Second, we analyze the effectiveness of every component of our unpaired method, \methodname, by incorporating them one at a time in Section~\ref{sec:ablation}. 
Finally, we show how our method works on paired settings and generates diverse outputs in Section~\ref{sec:exp_extension}. Please find the code, models, and interactive demos on our GitHub page \url{https://github.com/GaParmar/img2img-turbo}.

\myparagraph{Training details.}
Our total trainable parameters for the unpaired models on the driving datasets is 330 MB, including the LoRA weights,  zero-conv layer, and first conv layer of U-Net. 
Please find the hyperparameters and architecture details in Appendix~\ref{sup_sec:training_details}.

\myparagraph{Datasets.}
We conduct unpaired translation experiments on two commonly used datasets (Horse $\leftrightarrow$ Zebra and Yosemite Summer $\leftrightarrow$ Winter), and two higher resolution driving datasets (day $\leftrightarrow$ night and clear $\leftrightarrow$ foggy from BDD100k~\cite{yu2020bdd100k} and DENSE~\cite{Bijelic_2020_STF}). 
For the first two datasets, we follow CycleGAN~\cite{zhu2017unpaired} and load $286 \times 286$ images and use random $256 \times 256$ crops when training. During inference, we directly apply translation at $256 \times 256$. For driving datasets, we resize all images to $512 \times 512$ during both training and inference. For evaluation, 
we use the corresponding validation sets. %

\myparagraph{Evaluation Protocol.}
An effective image translation method must satisfy two key criteria: (1) matching the data distribution of the target domain and (2) preserving the structure of the input image in the translated output.
We evaluate the distribution matching using FID~\cite{heusel2017gans}, following the clean-FID's implementation~\cite{parmar2021cleanfid}. 
We assess adherence to the second criterion with DINO-Struct-Dist \cite{tumanyan2022splicing}, which measures the structure similarity of two images in feature space.  \emph{We report all DINO Structure scores multiplied by 100.}
A lower FID score indicates a closer match to the reference target distribution and greater realism, while a lower DINO-Struct-Dist suggests a more accurate preservation of the input structure in the translated image.
A low FID score with a high DINO-Struct-Dist indicates that a method is not able to adhere to the input structure. A low DINO-Struct-Dist but a high FID suggests that a method barely alters the input image. It is crucial to consider both of these scores together.
Additionally, we compare the inference runtime of all methods in Tables~\ref{tab:cmp_small_ds} and \ref{tab:cmp_driving_ds} on a Nvidia RTX A6000 GPU and include a human perceptual study.

\begin{table*}[t!]
    \centering
    \caption{
    \textbf{Evaluation on standard CycleGAN datasets (256 $\times$ 256).} 
    Comparison to prior GAN-based and Diffusion-based methods on standard CycleGAN datasets using FID to measure image quality and distribution alignment and DINO-Struct. to measure structure preservation. Our method achieves the lowest DINO-Struct. across all tasks and the lowest FID on all tasks except Horse $\rightarrow$ Zebra, while being magnitudes faster than diffusion-based models. Cycle-Diffusion obtains a slightly better FID but at the cost of large increase in DINO Struct., resulting in poor translation overall. 
 }
 \vspace{-5pt}
    \resizebox{\linewidth}{!}{
    \begin{tabular}{l c cc cc cc cc}
        \toprule 
        \multirow{3}{*}{\textbf{Method}} 
        & \multirow{3}{*}{\textbf{\shortstack[c]{Infrence \\ time }}} 
        & \multicolumn{2}{c}{\textbf{Horse $\rightarrow$ Zebra} }
        & \multicolumn{2}{c}{\textbf{Zebra $\rightarrow$ Horse} }
        & \multicolumn{2}{c}{\textbf{Summer $\rightarrow$ Winter} }
        & \multicolumn{2}{c}{\textbf{Winter $\rightarrow$ Summer} }
        \\

        \cmidrule(lr){3-4} \cmidrule(lr){5-6} \cmidrule(lr){7-8} \cmidrule(lr){9-10} 
        &
        & \multirow{2}{*}{\shortstack[c]{FID $\downarrow$ }}  
        & \multirow{2}{*}{\shortstack[c]{DINO \\ Struct. $\downarrow$ }} 

        & \multirow{2}{*}{\shortstack[c]{FID $\downarrow$ }}  
        & \multirow{2}{*}{\shortstack[c]{DINO \\ Struct. $\downarrow$ }} 

        & \multirow{2}{*}{\shortstack[c]{FID $\downarrow$ }}  
        & \multirow{2}{*}{\shortstack[c]{DINO \\ Struct. $\downarrow$ }} 

        & \multirow{2}{*}{\shortstack[c]{FID $\downarrow$ }}  
        & \multirow{2}{*}{\shortstack[c]{DINO \\ Struct. $\downarrow$ }} 
        
        \\ \\
        \cmidrule(lr){1-10}

        CycleGAN \cite{zhu2017unpaired}  & 0.01s
        & 74.9 & 3.2 
        & 133.8 & 2.6
        & 62.9 & 2.6
        & 66.1 & 2.3 
        \\

        CUT \cite{park2020contrastive} & 0.01s
        & 43.9 & 6.6 
        & 186.7 & 2.5 
        & 72.1 & 2.1 
        & 68.5 & 2.1 \\
        
        \hdashline

        SDEdit \cite{meng2022sdedit} & 1.56s
        & 77.2 & 4.0
        & 198.5 & 4.6  
        & 66.1 & 2.1 
        & 76.9 & 2.1\\
        
        Plug\&Play \cite{plug_and_play_Tumanyan} & 7.57s
        & 57.3 & 5.2  
        & 152.4 & 3.8 
        & 67.3 & 2.8 
        & 73.3 & 2.6\\
        
        Pix2Pix-Zero \cite{parmar2023zero}  & 14.75s
        & 81.5 & 8.0
        & 147.4 & 7.8
        & 68.0 & 3.0
        & 93.4 & 4.3 \\
        
        Cycle-Diffusion \cite{cyclediffusion} & 3.72s
        & \textbf{38.6} & 6.0 
        & 132.5 & 5.8 
        & 64.1 & 3.6
        & 70.3 & 3.6 \\

        DDIB \cite{su2022dual} & 4.37s
        & 44.4 & 13.1  
        & 163.3 & 11.1 
        & 90.8 & 7.2
        & 88.9 & 6.8 \\

        InstructPix2Pix \cite{brooks2022instructpix2pix} & 3.86s
        & 51.0 & 6.8 
        & 141.5 & 7.0 
        & 68.3 & 3.7 
        & 85.6 & 4.4 \\
        \hdashline

        \methodname & 0.13s
        & 41.0 & \textbf{2.1}
        & \textbf{127.5} & \textbf{1.8} 
        & \textbf{56.3} & \textbf{0.6} 
        & \textbf{60.7} & \textbf{0.6}\\

        \bottomrule 
    \end{tabular}
    }
    \vspace{-6pt}

    \label{tab:cmp_small_ds}
\end{table*}

\begin{table*}[t!]
    \centering
    \caption{
    \textbf{Comparison on 512 $\times$ 512  driving datasets.}  Our method outperforms all GAN-based and diffusion-based baselines on all driving datasets. InstructPix2pix gets a slightly lower DINO-Struct for Day $\rightarrow$ Night, but a much higher FID, thus not matching the target distribution well. Plug\&Play has similar results for Night $\rightarrow$ Day.  
 }
    \resizebox{\linewidth}{!}{
    \begin{tabular}{l c cc cc cc cc}
        \toprule 
        \multirow{3}{*}{\textbf{Method}} 
        & \multirow{3}{*}{\textbf{\shortstack[c]{Infrence \\ time }}} 
        & \multicolumn{2}{c}{\textbf{Day $\rightarrow$ Night} }
        & \multicolumn{2}{c}{\textbf{Night $\rightarrow$ Day} }
        & \multicolumn{2}{c}{\textbf{Clear $\rightarrow$ Foggy} }
        & \multicolumn{2}{c}{\textbf{Foggy $\rightarrow$ Clear} }
        \\

        \cmidrule(lr){3-4} \cmidrule(lr){5-6} \cmidrule(lr){7-8} \cmidrule(lr){9-10} 
        &
        & \multirow{2}{*}{\shortstack[c]{FID $\downarrow$ }}  
        & \multirow{2}{*}{\shortstack[c]{DINO \\ Struct. $\downarrow$ }} 

        & \multirow{2}{*}{\shortstack[c]{FID $\downarrow$ }}  
        & \multirow{2}{*}{\shortstack[c]{DINO \\ Struct. $\downarrow$ }} 

        & \multirow{2}{*}{\shortstack[c]{FID $\downarrow$ }}  
        & \multirow{2}{*}{\shortstack[c]{DINO \\ Struct. $\downarrow$ }} 

        & \multirow{2}{*}{\shortstack[c]{FID $\downarrow$ }}  
        & \multirow{2}{*}{\shortstack[c]{DINO \\ Struct. $\downarrow$ }} 
        
        \\ \\
        \cmidrule(lr){1-10}

        CycleGAN \cite{zhu2017unpaired} & 0.02s
        & 36.3 & 3.6 
        & 92.3 & 4.9 
        & 153.3 & 3.6 
        & 177.3 & 3.9 
        \\
        
        CUT \cite{park2020contrastive} & 0.03s
        & 40.7 & 3.5 
        & 98.5 & 3.8
        & 152.6 & 3.4 
        & 163.9 & 4.8  \\
        
        \hdashline

        SDEdit \cite{meng2022sdedit} & 3.10s
        & 111.7 & 3.4 
        & 116.1 & 4.1 
        & 185.3 & 3.1
        & 209.8 & 4.7\\

        Plug\&Play \cite{plug_and_play_Tumanyan} & 19.67s
        & 80.8 & 2.9 
        & 121.3 & \textbf{2.8} 
        & 179.6 & 3.6 
        & 193.5 & 3.5 \\
        
        Pix2Pix-Zero \cite{parmar2023zero}  & 43.28s
        & 81.3 & 4.7 
        & 188.6 & 5.8
        & 209.3 & 5.5
        & 367.2 & 13.0
        \\
        
        Cycle-Diffusion \cite{cyclediffusion}  & 11.38s
        & 101.1 & 3.1 
        & 110.7 & 3.7 
        & 178.1 & 3.6 
        & 185.8 & 3.1\\

        DDIB \cite{su2022dual} & 11.93s
        & 172.6 & 9.1
        & 190.5 & 7.8 
        & 257.0 & 13.0 
        & 286.0 & 7.2  \\

        InstructPix2Pix \cite{brooks2022instructpix2pix} & 11.41s
        & 80.7 & \textbf{2.1} 
        & 89.4 & 6.2 
        & 170.8 & 7.6 
        & 233.9 & 4.8 
        \\
        \hdashline

        \methodname & 0.29s
        & \textbf{31.3} & 3.0
        & \textbf{45.2} & 3.8 
        & \textbf{137.0} & \textbf{1.4}
        & \textbf{147.7} & \textbf{2.4} \\

        \bottomrule 
    \end{tabular}
    }
    \vspace{-18pt}

    \label{tab:cmp_driving_ds}
\end{table*}

\vspace{-10pt}
\subsection{Comparison to Unpaired Methods}\lblsec{baseline_comparison}
We compare \methodname to prior GAN-based unpaired image translation methods,  zero-shot image editing methods, and diffusion models trained for image editing using their publicly available code. Qualitatively, Figures~\ref{fig:cmp_unpaired_small_sd} and \ref{fig:cmp_unpaired_driving_ds} reveal that existing methods, both GAN-based and diffusion-based, struggle to achieve the right balance between output realism and structural preservation. 

\myparagraph{Comparison to GAN-based methods.}
We compare our method to two unpaired GAN models  - CycleGAN~\cite{zhu2017unpaired} and CUT~\cite{park2020contrastive}. We train these baseline models with default hyperparameters on all datasets for 100,000 steps and choose the best checkpoint. 
Tables~\ref{tab:cmp_small_ds} and \ref{tab:cmp_driving_ds} show quantitative comparisons on eight unpaired translation tasks. %
CycleGAN and CUT demonstrate effective performance, achieving low FID and DINO-Structure scores on simpler, object-centric datasets, such as horse $\rightarrow$ zebra (Figure~\ref{fig:sup_baseline_h2z_gan}). 
Our method slightly outperforms these in terms of both FID and DINO-structure distance metrics. 
However, for more complex scenes, such as night $\rightarrow$ day, CycleGAN and CUT get significantly higher FID scores than our method, often hallucinating undesirable artifacts (Figure~\ref{fig:sup_baseline_day2night_gan}).

\myparagraph{Comparison to diffusion-based editing methods.}
Next, we compare our method to several diffusion-based methods in Tables~\ref{tab:cmp_small_ds} and \ref{tab:cmp_driving_ds}. First, we consider recent zero-shot image translation methods, including SDEdit~\cite{meng2022sdedit}, Plug-and-Play~\cite{plug_and_play_Tumanyan}, pix2pix-zero~\cite{parmar2023zero}, CycleDiffusion~\cite{cyclediffusion}, and DDIB~\cite{su2022dual} that use a pre-trained text-to-image diffusion model and translate the images through different text prompts.  Note that the original DDIB implementation involves training two separate domain-specific diffusion models from scratch. To improve its performance and have a fair comparison, we replace the domain-specific models with a pre-trained text-to-image model. We also compare to Instruct-pix2pix~\cite{brooks2022instructpix2pix}, a conditional diffusion model trained for text-based image editing.

As shown in Table~\ref{tab:cmp_small_ds} and Figure~\ref{fig:sup_baseline_h2z_diffusion},  on object-centric datasets such as a horse $\rightarrow$ zebra, these methods can generate realistic zebras but struggle to precisely match the object poses, as indicated by consistently large DINO-structure scores. 
On driving datasets, those editing methods perform noticeably worse due to three reasons: (1) the models struggle to generate complex scenes containing multiple objects, (2) these methods (except Instruct-pix2pix) need to first invert the images to a noise map, introducing potential artifacts, and (3) the pre-trained models cannot synthesize street view images similar to the one captured by the driving datasets. 
Table~\ref{tab:cmp_driving_ds} and Figure~\ref{fig:sup_baseline_day2night_diffusion} show that across all four driving translation tasks, these methods output poor quality images, reflected by a high FID score, and do not adhere to input image structure, reflected in high DINO-Structure distance values.

\begin{table*}[t!]
    \centering
    \caption{
    \textbf{Human Preference Evaluation.}  We conduct a study that asks users to pick images that look more like the target domain. We rate every image in the validation with 3 different users. Our method is preferred across all datasets, with the exception of Clear to Foggy. 
    }
    \begin{tabular}{l cc cc }
        \toprule 
        \textbf{Method} & \textbf{Day $\rightarrow$ Night}  & \textbf{Night $\rightarrow$ Day} 
        & \textbf{Clear $\rightarrow$ Foggy}  & \textbf{Foggy $\rightarrow$ Clear} 
        \\ 
        \cmidrule(lr){1-5}
        CycleGAN \cite{zhu2017unpaired} &
        45.9\%     & 37.4\% &
        45.4\%     & 26.7\%\\
        
        \textbf{ours} & 
        \textbf{54.1\%} & \textbf{62.6\%} &
        \textbf{54.6\%} & \textbf{73.3\%}\\
        
        \hdashline
        InstructPix2Pix \cite{brooks2022instructpix2pix} & 
        25.1\% & 29.1\% &
        \textbf{69.4\%} & 13.3\% \\ 
        
        \textbf{ours}  &
        \textbf{74.9\%} & \textbf{70.9\%} &
        30.6\% & \textbf{86.7\%}\\
        \bottomrule 
    \end{tabular}

    \label{tab:human_eval}
\end{table*}

\myparagraph{Human Preference Study}
Next, we conduct a human preference study on Amazon Mechanical Turk (AMT) to evaluate the quality of images produced by the different methods. We use the complete validation set from the relevant datasets, with each comparison independently evaluated by three unique users. 
We present the outputs of two models side-by-side and ask users to choose which one follows the target prompt more accurately in an unlimited time.
For instance, we collect 1,500 comparisons for the Day to Night translation task with 500 validation images. %
The\ prompt presented to the users is: 
``Which image looks more like a real picture of a driving scene taken in the night?''

Table~\ref{tab:human_eval} compares our method to CycleGAN~\cite{zhu2017unpaired}, the best performing GAN-based method, and Instruct-Pix2Pix~\cite{brooks2022instructpix2pix}, the best performing diffusion-based method. 
Our method outperforms the two baselines across all datasets, except for the Clear to Foggy translation task. In this case, users favor InstructPix2Pix's results, as it outputs more artistic fog images. However, InstructPix2Pix fails to preserve the input structure, as indicated by its high DINO-Struct score (7.6) compared to ours (1.4). Moreover, its results substantially diverge from the target fog dataset, reflected by a high FID score (170.8) compared to ours (137.0), as noted in Table~\ref{tab:cmp_driving_ds}.

\begin{table*}[t!]
    \caption{
    \textbf{Ablation with Horse to Zebra.} 
    The values in parentheses reflect the relative change compared to our final method. First, Conf. A trains the unpaired translation model with randomly initialized weights and suffers from a large FID increase. Next, Conf. B, C, and D try different input types and show that direct input achieves the best performance. Finally, our method adds skip connections to Conf. D and shows an improvement in structure preservation. 
    Ablation on other tasks is shown in Appendix~\ref{sup_sec:additional_ablations}. 
    }
    \centering
    \resizebox{\linewidth}{!}{
    \begin{tabular}{l c c c | cc cc cc cc}
        \toprule 
        \multirow{3}{*}{\textbf{Method} }
        & \multirow{3}{*}{\shortstack[c]{\textbf{Input} \\ \textbf{Type} }}  
        & \multirow{3}{*}{\shortstack[c]{\textbf{Skip} }}  
        & \multirow{3}{*}{\shortstack[c]{\textbf{Pre} \\ \textbf{-trained} }}  
        & \multicolumn{2}{c}{\textbf{Horse $\rightarrow$ Zebra} }
        & \multicolumn{2}{c}{\textbf{Zebra $\rightarrow$ Horse} }
        \\

        \cmidrule(lr){5-6} \cmidrule(lr){7-8} \cmidrule(lr){9-10} \cmidrule(lr){11-12} 

        &&&& \multirow{2}{*}{\shortstack[c]{FID $\downarrow$ }}  
        & \multirow{2}{*}{\shortstack[c]{DINO \\ Struct. $\downarrow$ }} 

        & \multirow{2}{*}{\shortstack[c]{FID $\downarrow$ }}  
        & \multirow{2}{*}{\shortstack[c]{DINO \\ Struct. $\downarrow$ }} 
        
        \\ \\
        \cmidrule(lr){1-12}
        Conf. A & Direct Input & x & x
        & 128.6 (+214\%)  & 5.2 (+148\%)
        & 167.1 (+31\%)  & 4.6 (+156\%)
        \\
        
        Conf. B & ControlNet & x & \checkmark 
        & 41.2 (+0\%)  & 7.3 (+248\%)
        & \textbf{99.4 (-22\%)}  & 8.6 (+378\%)\\

        Conf. C & T2I-Adapter & x & \checkmark 
        & 55.4 (+35\%)  & 4.7 (+124\%)
        & 135.4 (+6\%) & 4.8 (+167\%)\\

        Conf. D & Direct Input & x & \checkmark
        & \textbf{40.1 (-2\%)}  & 4.4 (+110\%)
        & 116.2 (-9\%) & 3.0 (+67\%) \\

        \hdashline
        Ours & Direct Input & \checkmark & \checkmark
        & 41.0 & \textbf{2.1}
        & 127.5 & \textbf{1.8}  \\

        \bottomrule 
    \end{tabular}
    }
\vspace{-10pt}
    \label{tab:ablation_study}
\end{table*}

\begin{figure}[t!]
    \centering
    \includegraphics[width=0.94\linewidth]{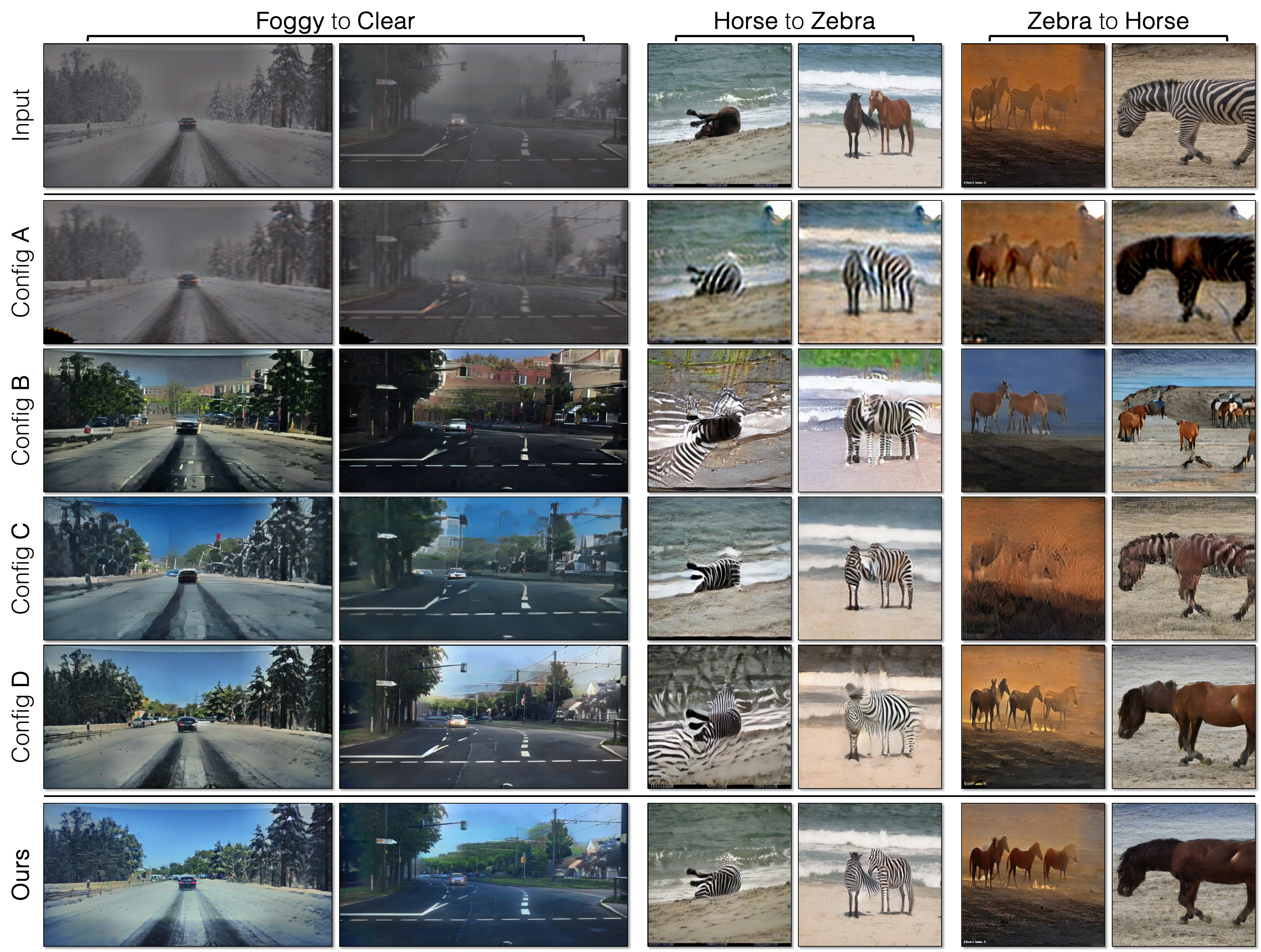}
        \vspace{-8pt}

    \caption{{\bf Ablating individual components.} Our final formulation achieves the best content preservation and realism, compared to other design choices described in Table~\ref{tab:ablation_study}.}
        \vspace{-15pt}

    \label{fig:abl_all}
\end{figure}
\vspace{-10pt}
\subsection{Ablation Study}\label{sec:ablation}
Here, we show the effectiveness of our algorithmic designs through an extensive ablation study in Table~\ref{tab:ablation_study} and Figure~\ref{fig:abl_all}.  %

\myparagraph{Using pre-trained weights.}
First, we assess the impact of using a pre-trained network. In Table~\ref{tab:ablation_study} Config A, we train an unpaired model on the Horse $\leftrightarrow$ Zebra dataset but
with randomly initialized weights rather than pre-trained weights. 
Without leveraging the prior from the pre-trained text-to-image model, the output images look unnatural, as shown in Figure~\ref{fig:abl_all} Config A. 
This observation is corroborated by a large increase in FID across both tasks in Table~\ref{tab:ablation_study}.

\myparagraph{Different ways of adding conditioning inputs.}
Next, we compare three ways of adding structure input to the model. Config B uses a ControlNet Encoder~\cite{zhang2023adding}, Config C uses the T2I-Adapter \cite{mou2023t2i}, and finally, Config D directly feeds the input image to the base network without any additional branches.  
Config B obtains a comparable FID to Config D. However, it also has a significantly higher DINO-Structure distance, indicating that the ControlNet encoder struggles to match the input's structure. 
This is also observed in Figure~\ref{fig:abl_all}; Config B (third row) consistently changes the scene structure and hallucinates new objects, such as partial buildings in the case of driving scenes and unnatural zebra patterns for the horse-to-zebra translation. 
Config C uses a lightweight T2I-Adapter to learn the structure and achieves worse FID and DINO-Struct scores, and output images that have several artifacts and poor structure preservation. 

\myparagraph{Skip Connections and trainable encoder and decoder.}
Finally, we can see the effects of skip connections by comparing Config D to our final method $\methodname$ in Table~\ref{tab:ablation_study} and Figure~\ref{fig:abl_all}. Across all tasks, adding skip connections and training the encoder and decoder jointly can significantly improve structure preservation, albeit at the cost of a small increase in FID.

\myparagraph{Additional results.} 
Please see Appendix~\ref{sup_sec:additional_ablations} and \ref{sup_sec:analysis} for additional ablation studies on other datasets, the effect of model training with varying numbers of training images, and the role of encoder-decoder fine-tuning.

\begin{figure}[t!]
    \centering
    \includegraphics[width=0.9\linewidth]{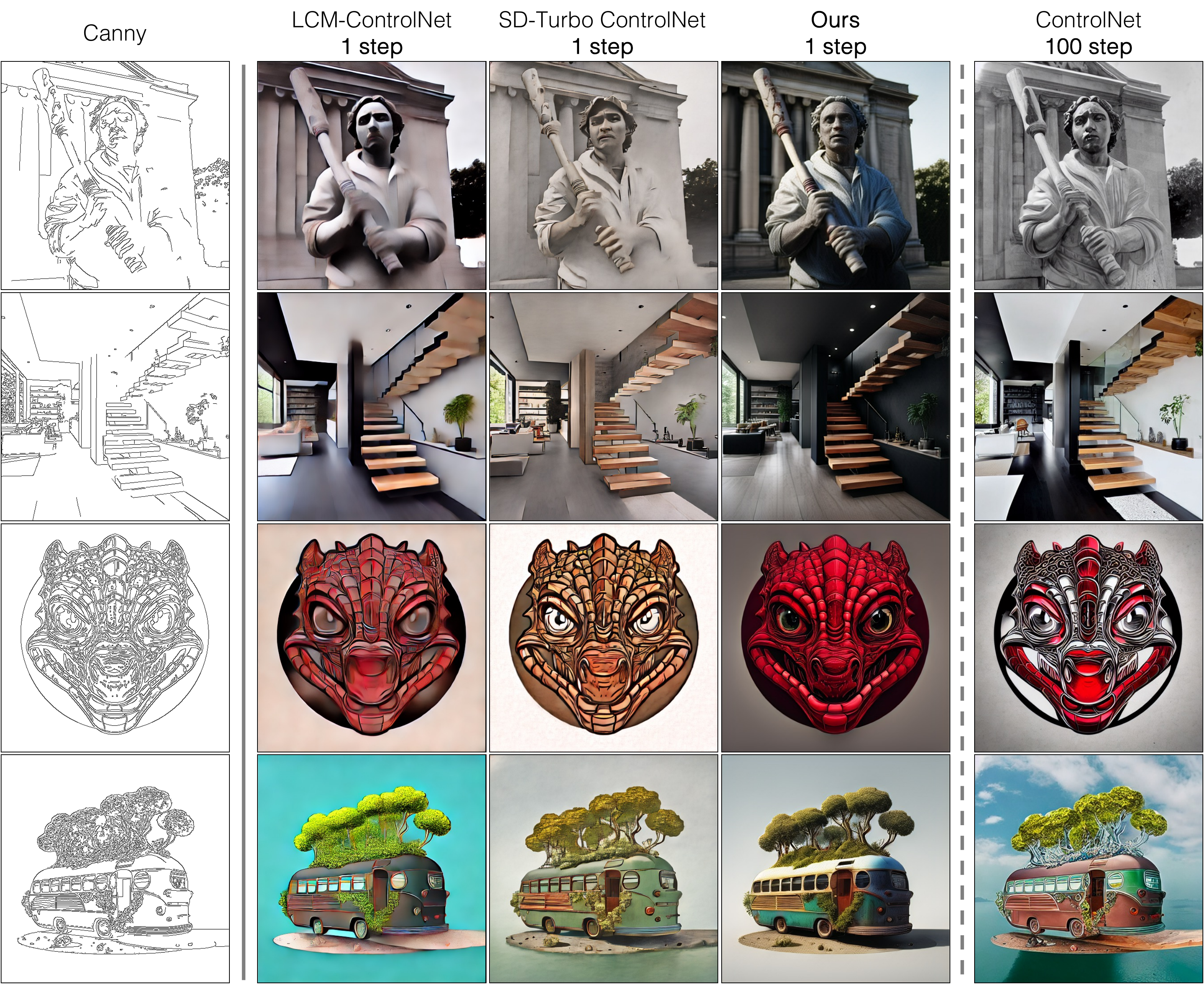}
     \vspace{-8pt}
    \caption{{\bf Comparison on paired edge-to-image task (512 $\times$ 512).} Our method (runtime: 0.29s) achieves higher realism than existing one-step methods and is competitive with the 100-step ControlNet (runtime: 18.85s).}
    \label{fig:cmp_paired}
     \vspace{-18pt}
\end{figure}
\vspace{-10pt}
\subsection{Extensions}
\vspace{-5pt}
\label{sec:exp_extension}
\myparagraph{Paired translation.} %
We train Edge2Photo and Sketch2Photo models on a community-collected dataset of 300K artistic images~\cite{edgedataset}.
We extract Canny edges~\cite{canny1986computational} and HED contours~\cite{xie2015holistically}. 
As our method and baselines use different datasets, we show visual comparisons instead of conducting FID evaluation. 
More details on training data and preprocessing are included in Appendix~\ref{sup_sec:training_details}.

We compare our paired method \pairedname to existing one-step and multi-step translation methods in Figure~\ref{fig:cmp_paired}, including 
two one-step baselines that use Latent Consistency Models ~\cite{luo2023latent} and the Stable Diffusion - Turbo ~\cite{sauer2023adversarial} with a ControlNet adapter. While these approaches can produce results in one step, their image quality degrades. Next, we compare it to the vanilla ControlNet, which uses Stable Diffusion with 100 steps. We additionally use classifier-free guidance and a long descriptive negative prompt for the 100-step ControlNet baseline. 
This approach can generate more pleasing outputs compared to the one-step baselines, as shown in Figure~\ref{fig:cmp_paired}. Our method generates compelling outputs with only one forward pass, without negative prompting or classifier-free guidance.

\begin{figure}[t!]
    \centering
    \includegraphics[width=0.98\linewidth]{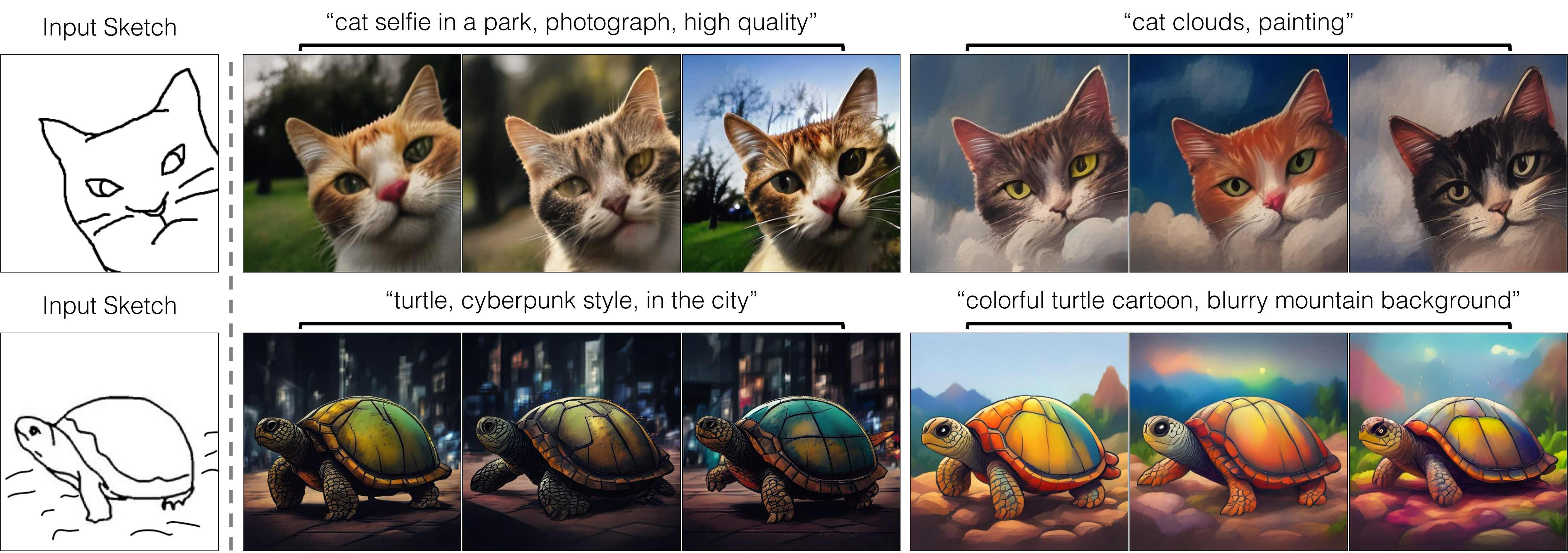}
     \vspace{-10pt}
    \caption{{\bf Generating diverse outputs.} By varying the input noise map, our method can generate diverse outputs from the same input conditioning. Moreover, the output style can be controlled by changing the text conditioning.}
    \label{fig:diversity_paired}
     \vspace{-15pt}
\end{figure}
\myparagraph{Generating diverse outputs.}\lblsec{exp_multimodal}
Finally, in Figure~\ref{fig:diversity_paired}, we show that our method can be used to generate diverse outputs as described in Section~\ref{sec:extension}. 
Given the same input sketch and user prompt, we can sample different noise maps and generate diverse multi-modal outputs, such as cats in different styles, variations in the background, and turtles with different shell patterns.

\vspace{-10pt}
\section{Discussion and Limitations}
\vspace{-8pt}
 \label{sec:discussion}
Our work suggests that one-step pre-trained models can serve as a strong and versatile backbone model for many downstream image synthesis tasks. Adapting these models to new tasks and domains can be achieved through various GANs objectives,  without the need for multi-step diffusion training. Our model training only requires a small number of additional trainable parameters. %

 \myparagraph{Limitations.}
 Although our model can produce visually appealing results with a single step, it does have limitations.  First,  we cannot specify the strength of the guidance, as our backbone model SD-Turbo does not use classifier-free guidance. Guided distillation~\cite{meng2023distillation} could be a promising solution to enable guidance control.  Second, our method does not support negative prompt, a convenient way of reducing artifacts. Third, model training with cycle-consistency loss and high-capacity generators is memory-intensive. Exploring one-sided method~\cite{park2020contrastive} for higher-resolution image synthesis is a meaningful next step.

 \myparagraph{Acknowledgments.} We thank Anurag Ghosh, Nupur Kumari, Sheng-Yu Wang, Muyang Li, Sean Liu, Or Patashnik, George Cazenavette, Phillip Isola, and Alyosha Efros for fruitful discussions and valuable feedback on our manuscript. %
 This work was partly supported by GM Research Israel, NSF IIS-2239076, the Packard Fellowship, and Adobe Research.

{\small
\bibliographystyle{splncs04}
\bibliography{main}
}

\clearpage

\appendix
\label{sup_sec:overview}

\noindent{\Large\bf Appendix}
\vspace{5pt}

Next, we start with Section~\ref{sup_sec:additional_ablations}, which provides additional ablation study results on more datasets. 
Section~\ref{sup_sec:additional_baseines} follows with more baseline comparisons with all GAN-based and Diffusion-based baselines. 
Section~\ref{sup_sec:analysis} shows an additional analysis of the Condition Encoder conflict, the effects of varying the dataset size, and the role of encoder-decoder finetuning. 
Finally, in Section~\ref{sup_sec:training_details}, we provide the hyperparameters and training details.

\section{Additional Ablation Study} \label{sup_sec:additional_ablations}
\begin{figure}
    \centering
    \includegraphics[width=\linewidth]{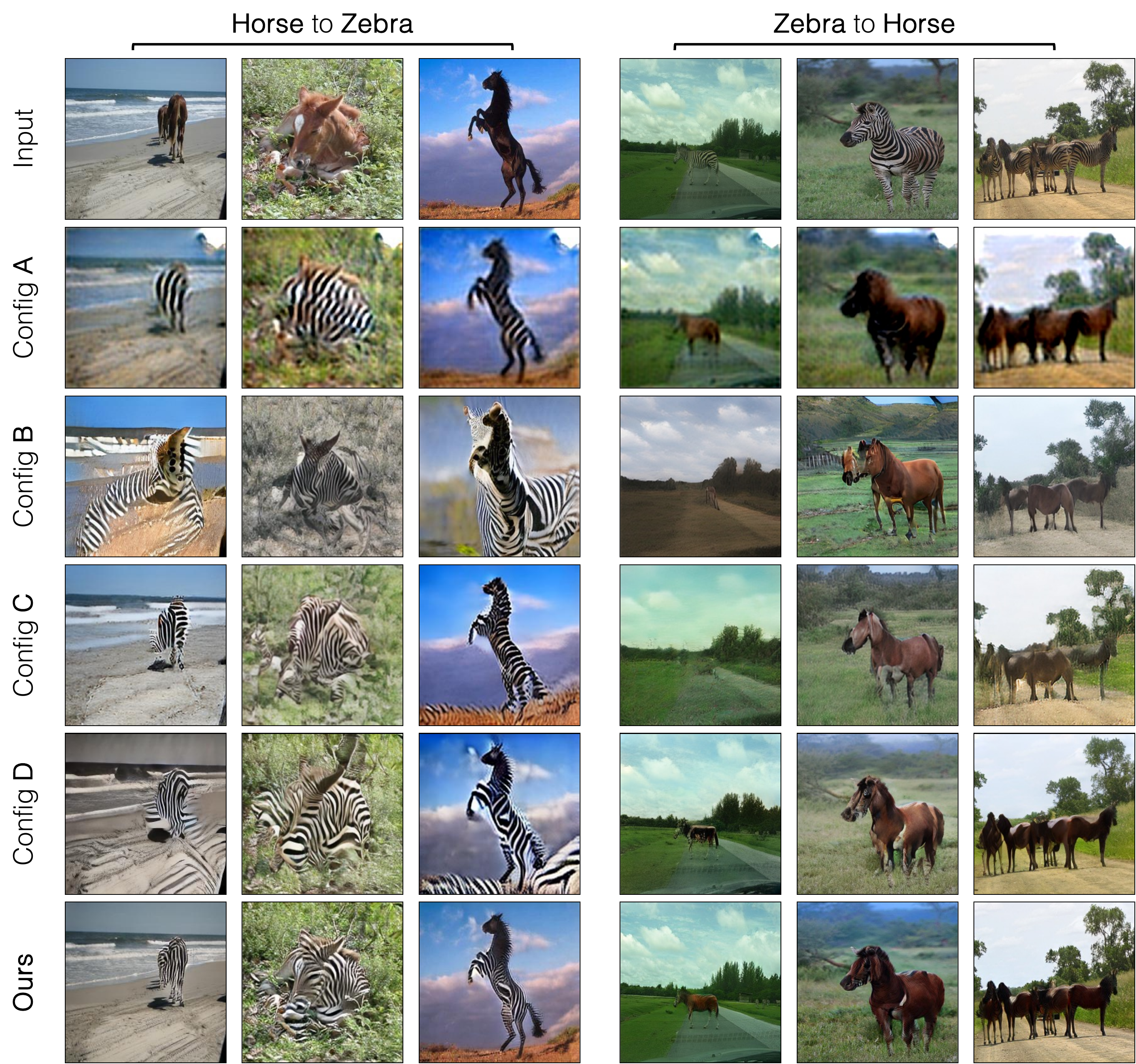}

    \caption{{\bf Ablating individual components.} Additional ablation results on the Horse $\leftrightarrow$ Zebra dataset. Our final method, shown in the bottom row, achieves the best translation results.}

    \label{fig:sup_abl_horse_zebra}
\end{figure}

\begin{figure}
    \centering
    \includegraphics[width=\linewidth]{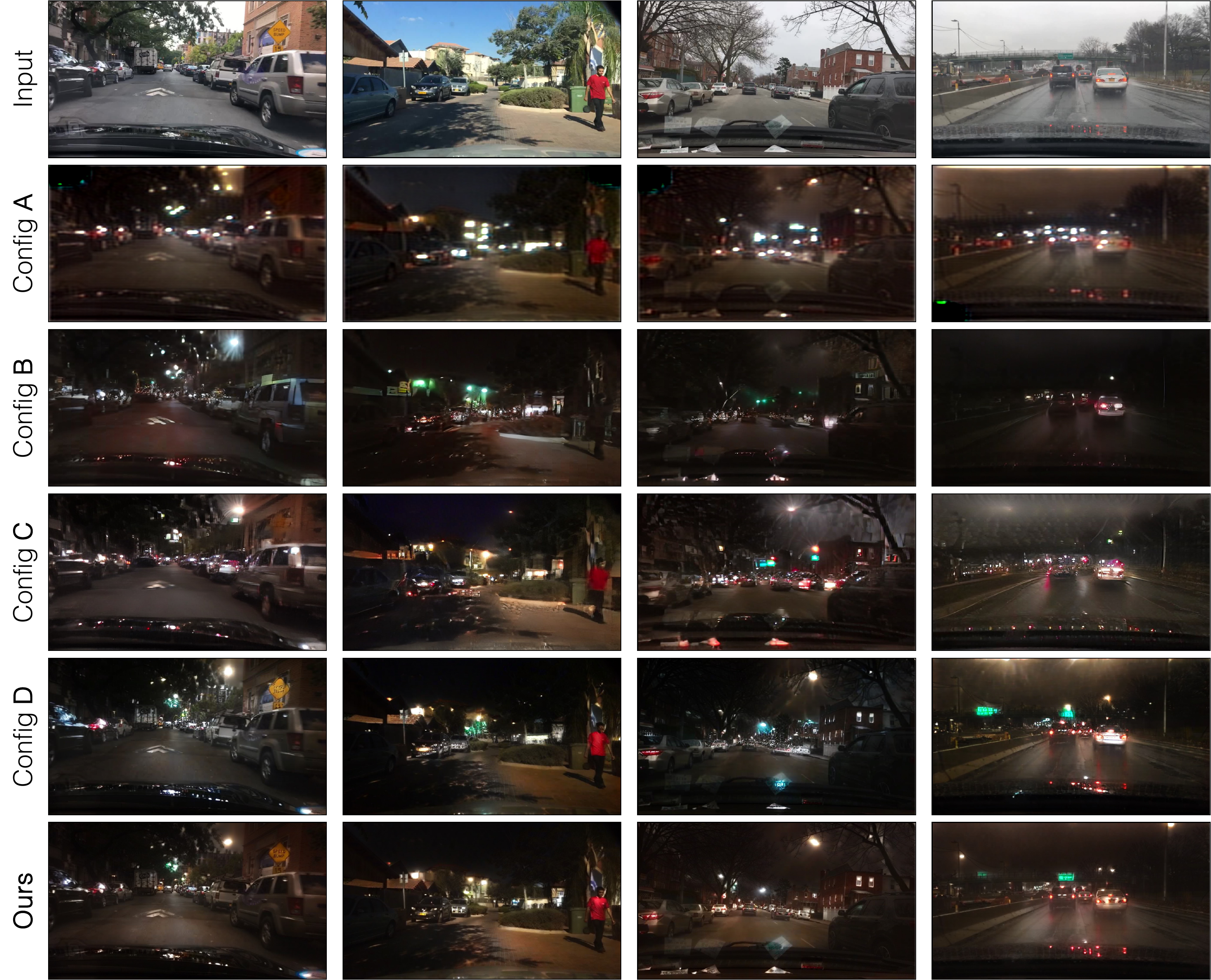}

    \caption{{\bf Ablating individual components.} Additional ablation results on the Day $\rightarrow$ Night translation. Our method, shown in the bottom row, generates the most convincing translations with the best detail preservation. Please zoom in to see the differences. }

    \label{fig:sup_abl_day_night}
\end{figure}

\begin{figure}
    \centering
    \includegraphics[width=\linewidth]{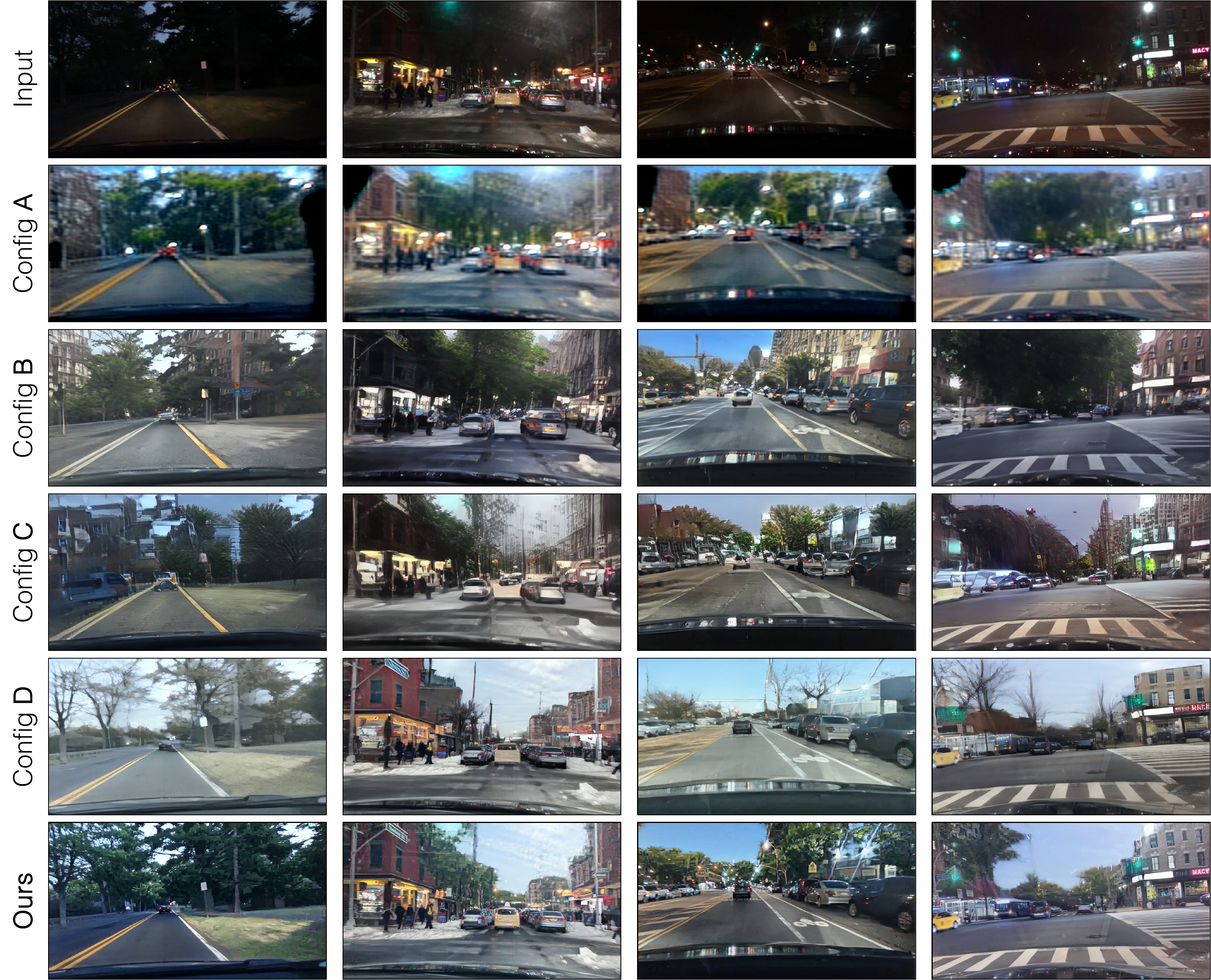}

    \caption{{\bf Ablating individual components.} Additional ablation results on the Night $\rightarrow$ Day translation. Our method, shown in the bottom row, generates the most convincing translations with the best detail preservation. Please zoom in to see the differences.}

    \label{fig:sup_abl_night_day}
\end{figure}

Table 3 in the main paper shows the results of an ablation study on the Horse to Zebra translation. 
We show more qualitative ablation results on this dataset in Figure~\ref{fig:sup_abl_horse_zebra}. 
Next, we perform the same ablation on the Day to Night translation qualitatively in Figures~\ref{fig:sup_abl_day_night},~\ref{fig:sup_abl_night_day} and Table~\ref{tab:sup_abl_day_night}. Similar to the main paper, we compare to four variants: (1) Config A uses randomly initialized weights rather than pre-trained weights, (2)  Config B uses a ControlNet Encoder~\cite{zhang2023adding}, (3) Config C uses the T2I-Adapter \cite{mou2023t2i}, and (4) Config D directly feeds the input image to the base network without skip connections.

Our full method outperforms all other variants in terms of distribution matching (FID) and structure preservation (DINO Structure Distance).

\section{Additional Baseline Comparisons} \label{sup_sec:additional_baseines}
\begin{figure}[t!]
    \centering
    \includegraphics[width=\linewidth]{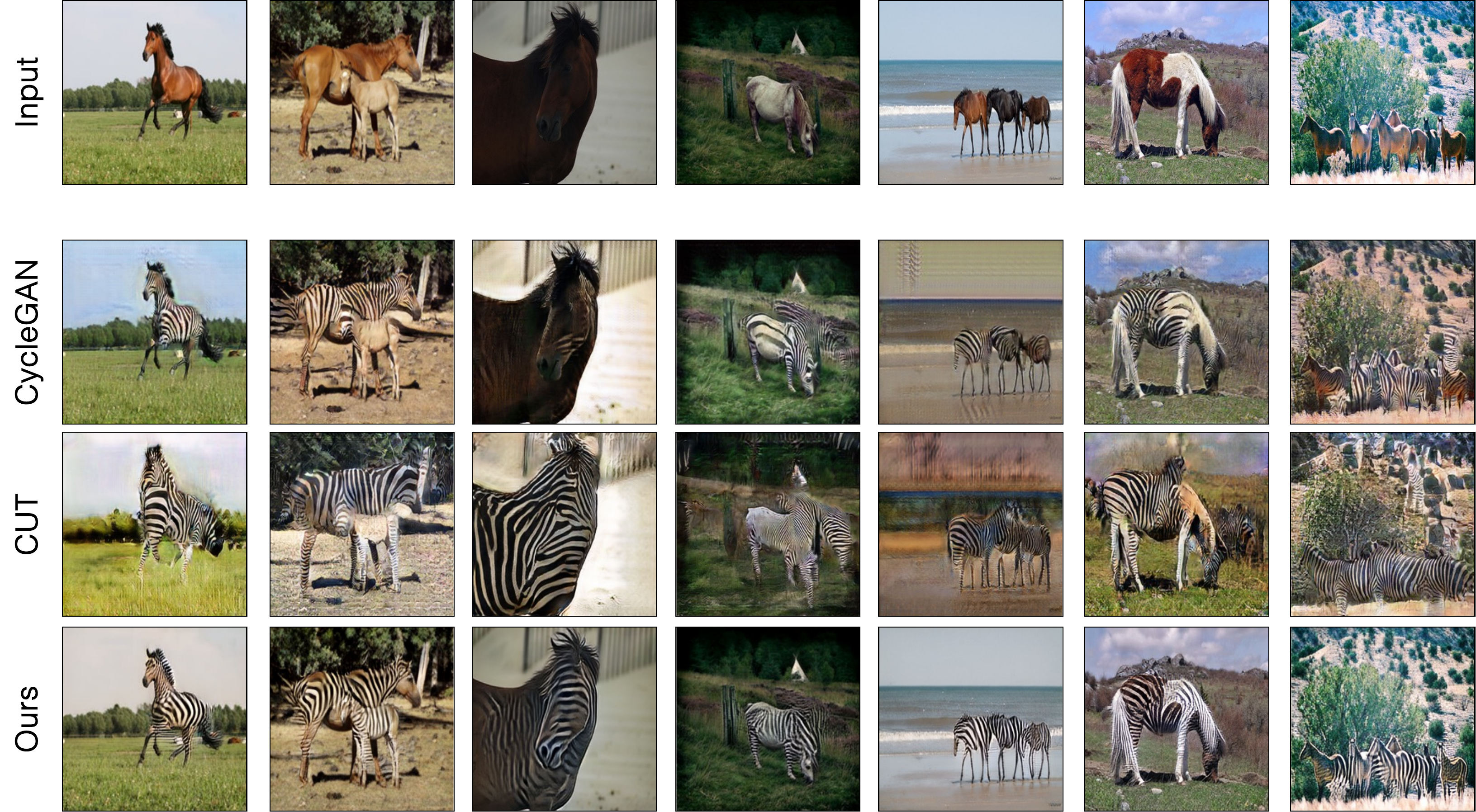}
    \caption{{\bf Comparison to GAN-based baselines.} Additional comparison to CycleGAN and CUT on Horse $\leftrightarrow$ Zebra translation task.}
    \label{fig:sup_baseline_h2z_gan}
\end{figure}

\begin{figure}[t!]
    \centering
    \includegraphics[width=\linewidth]{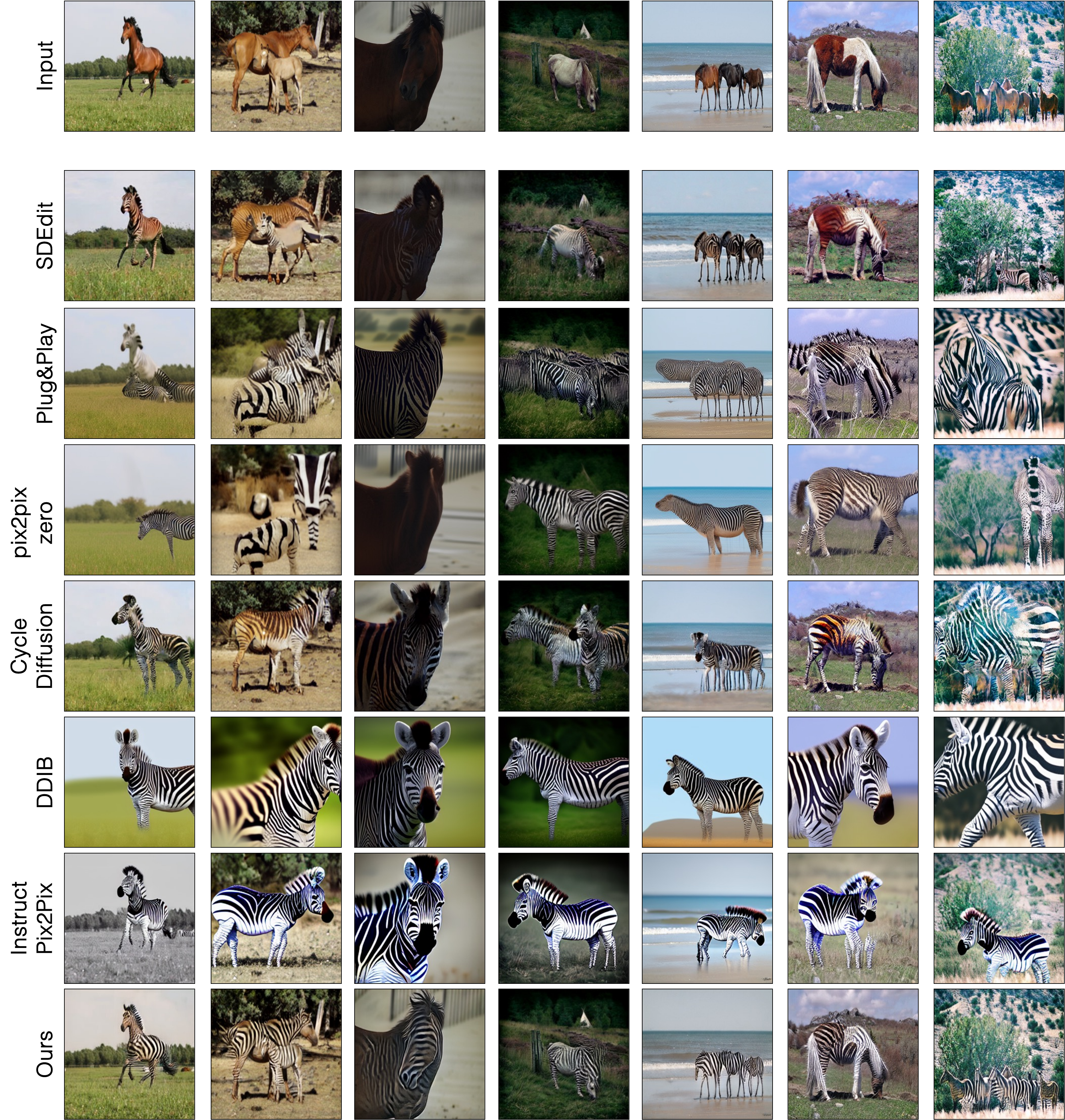}
    \caption{{\bf Comparison to Diffusion-based baselines.} Additional comparison to diffusion-based baselines on Horse $\leftrightarrow$ Zebra translation task.}
    \label{fig:sup_baseline_h2z_diffusion}
\end{figure}

\begin{figure}[t!]
    \centering
    \includegraphics[width=\linewidth]{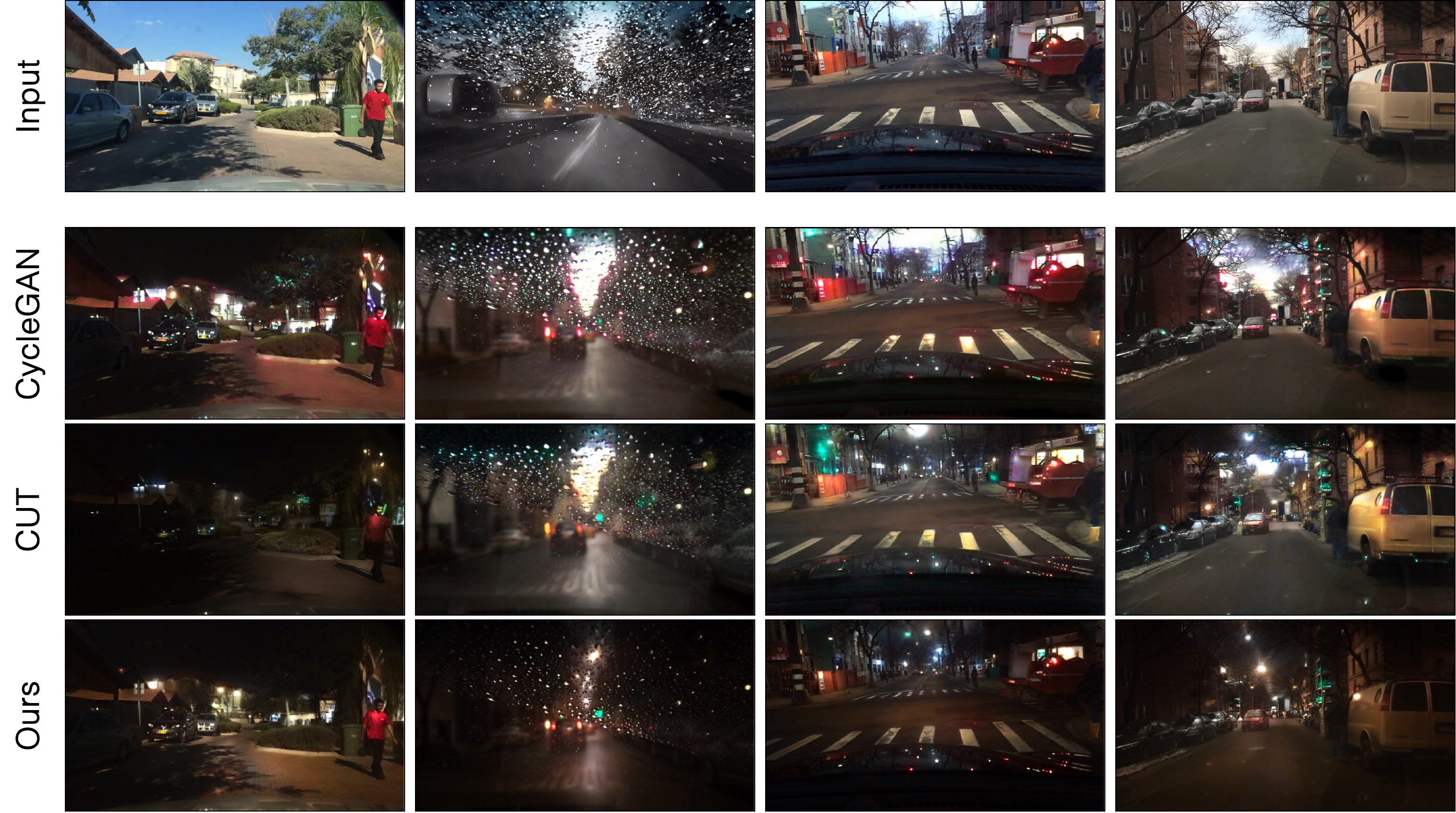}
    \caption{{\bf Comparison to GAN-based baselines.} Additional comparison to CycleGAN and CUT on the Day $\rightarrow$ Night translation task.}
    \label{fig:sup_baseline_day2night_gan}
\end{figure}

\begin{figure}[t!]
    \centering
    \includegraphics[width=\linewidth]{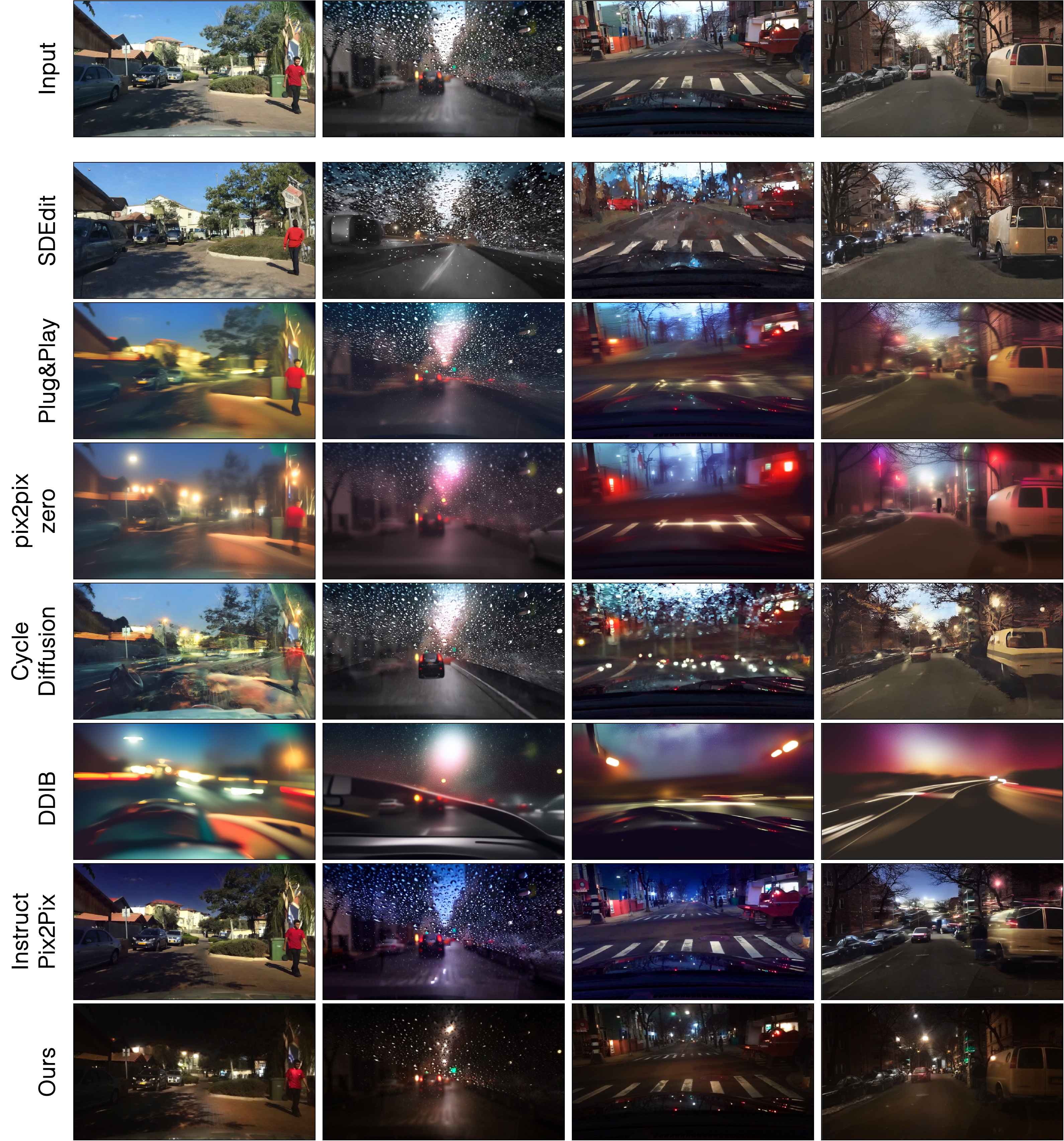}
    \caption{{\bf Comparison to Diffusion-based baselines.} Additional comparison to several diffusion-based baselines on the Day $\rightarrow$ Night translation task.}
    \label{fig:sup_baseline_day2night_diffusion}
    \vspace{-10pt}
\end{figure}

Figures 5 and 6 in the main paper show a comparison of our method with the best-performing GAN baseline and the best-performing diffusion-based baseline. Here, we show additional qualitative comparisons with all GAN baselines in Figures~\ref{fig:sup_baseline_h2z_gan} and \ref{fig:sup_baseline_day2night_gan}, as well as all diffusion-based baselines in Figures~\ref{fig:sup_baseline_h2z_diffusion} and \ref{fig:sup_baseline_day2night_diffusion}. Our method consistently produces more realistic outputs while retaining the structure of input images. %


\begin{table*}[t!]
    \caption{
    \textbf{Ablation with Day to Night.} 
    The values in parentheses reflect the relative change compared to our final method. First, Conf. A trains the unpaired translation model with randomly initialized weights and suffers from a large FID increase. Next, Conf. B, C, and D try different input types and show that direct input achieves the best performance. Finally, our method adds skip connections to Conf. D and shows an improvement in both distribution matching and structure preservation.
    }
    \centering
    \resizebox{\linewidth}{!}{
    \begin{tabular}{l c c c | cc cc cc cc}
        \toprule 
        \multirow{3}{*}{\textbf{Method} }
        & \multirow{3}{*}{\shortstack[c]{\textbf{Input} \\ \textbf{Type} }}  
        & \multirow{3}{*}{\shortstack[c]{\textbf{Skip} }}  
        & \multirow{3}{*}{\shortstack[c]{\textbf{Pre} \\ \textbf{-trained} }}  
        & \multicolumn{2}{c}{\textbf{Day $\rightarrow$ Night} }
        & \multicolumn{2}{c}{\textbf{Night $\rightarrow$ Day} }
        \\

        \cmidrule(lr){5-6} \cmidrule(lr){7-8} \cmidrule(lr){9-10} \cmidrule(lr){11-12} 

        &&&& \multirow{2}{*}{\shortstack[c]{FID $\downarrow$ }}  
        & \multirow{2}{*}{\shortstack[c]{DINO \\ Struct. $\downarrow$ }} 

        & \multirow{2}{*}{\shortstack[c]{FID $\downarrow$ }}  
        & \multirow{2}{*}{\shortstack[c]{DINO \\ Struct. $\downarrow$ }} 
        
        \\ \\
        \cmidrule(lr){1-12}
        Conf. A & Direct Input & x & x
        & 86.3 (+176\%)  & 4.4 (+47\%)
        & 105.8 (+134\%)  & 5.3 (+39\%)
        \\
        
        Conf. B & ControlNet & x & \checkmark 
        & 35.8 (+14\%)  & 5.4 (+80\%)
        & 48.7 (+8\%)   & 5.5 (+45\%) \\

        Conf. C & T2I-Adapter & x & \checkmark 
        & 34.2 (+9\%)  & 4.2 (+40\%)
        & 54.6 (+21\%)  & 6.4 (+68\%)\\

        Conf. D & Direct Input & x & \checkmark
        & 33.5 (+7\%)  & 4.0 (+33\%)
        & 48.5 (+7\%)  & 4.9 (+29\%) \\

        \hdashline
        Ours & Direct Input & \checkmark & \checkmark
        & \textbf{31.3} & \textbf{3.0}
        & \textbf{45.2} & \textbf{3.8}  \\

        \bottomrule 
    \end{tabular}
    }
\vspace{-10pt}
    \label{tab:sup_abl_day_night}
\end{table*}


\section{Additional Analysis} \label{sup_sec:analysis}
\begin{figure}[t!]
    \centering
    \includegraphics[width=\linewidth]{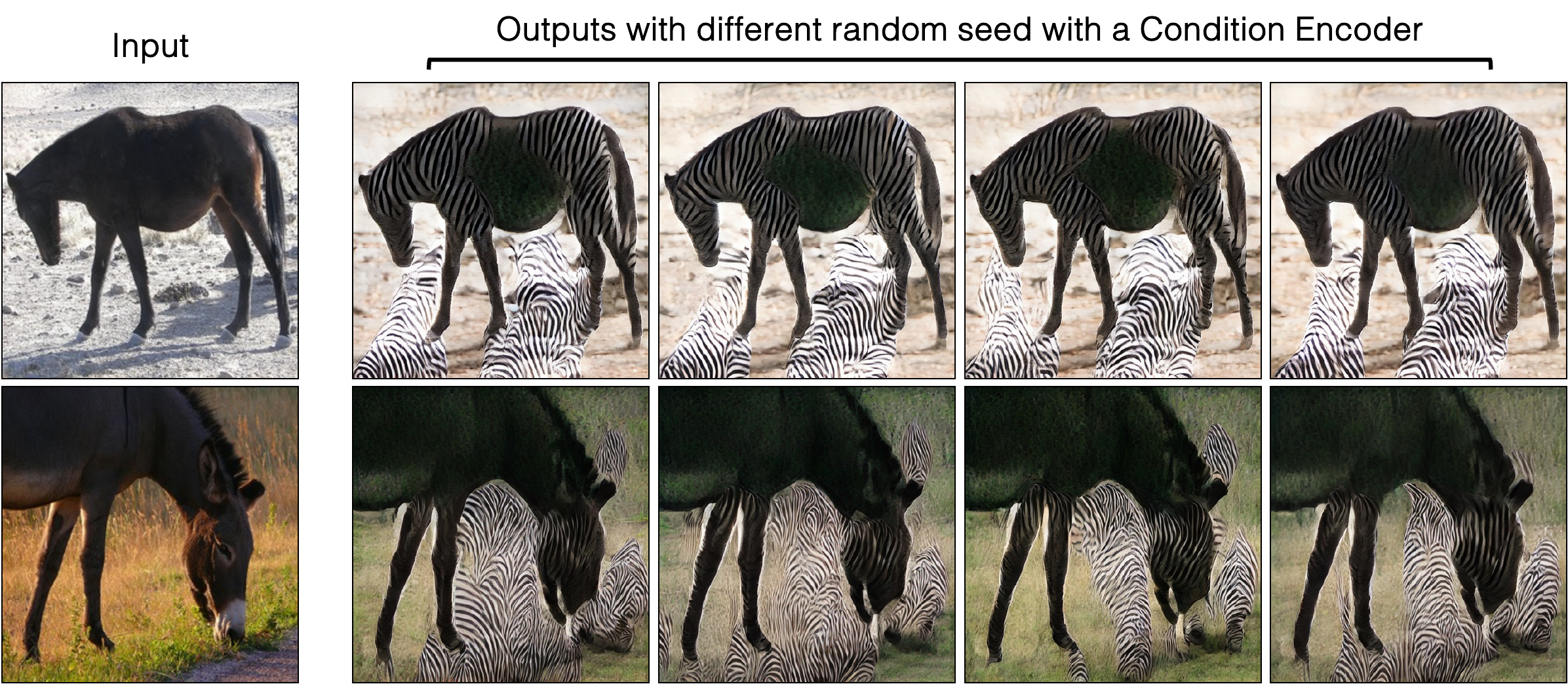}
    \caption{{\bf Different outputs with the same input image and different noise maps.} We observe that the noise maps do not alter the image structure, suggesting that the noises have been largely ignored. }
    \label{fig:sup_noise_ignored}
\end{figure}

\myparagraph{Conflict with Condition Encoder.} 
Figure 3 in the main paper illustrates the conflicting features when a conditioning image is added through a separate encoder. Here, we show that using a Condition Encoder, as depicted in Figure~\ref{fig:conditioning_conflict} of the main paper, results in the original network getting ignored. In Figure~\ref{fig:sup_noise_ignored}, we show the output with different noise maps but the same condition image. The different noise maps generate perceptually similar output images, indicating that the original SD-Turbo Encoder features have been ignored.

\begin{table*}[t!]
    \centering
    \caption{
    \textbf{Training with a different number of input images.}   
    }
    \begin{tabular}{cc | cc cc}
        \toprule 
        \multirow{3}{*}{\shortstack[c]{\textbf{\# Day} \\ \textbf{Image} }}  &
        \multirow{3}{*}{\shortstack[c]{\textbf{\# Night} \\ \textbf{Image} }} & 
        \multicolumn{2}{c}{\textbf{Day $\rightarrow$ Night} } & 
        \multicolumn{2}{c}{\textbf{Night $\rightarrow$ Day} } \\
        
        \cmidrule(lr){3-4} \cmidrule(lr){5-6}
        
        & & \multirow{2}{*}{\shortstack[c]{FID \\ $\downarrow$ }}   & 
        \multirow{2}{*}{\shortstack[c]{DINO \\ Struct. $\downarrow$ }}
        & \multirow{2}{*}{\shortstack[c]{FID \\ $\downarrow$ }}   & 
        \multirow{2}{*}{\shortstack[c]{DINO \\ Struct. $\downarrow$ }}\\ 
        \\
        \cmidrule(lr){1-6}
        10     & 10     &
        42.4 & 3.0 &
        65.6 & 4.0 \\
        
        100    & 100    & 
        31.8 & 3.3 &
        47.4 & 3.8  \\

        1000   & 1000   &
        31.2 & 3.4 &
        47.4 & 3.8 \\

        36,728 & 27,971 &
        31.3 & 3.0 &
        45.2 & 3.8 \\
        \bottomrule 
    \end{tabular}

    \label{tab:sup_num_training_images}
\end{table*}

\myparagraph{Varying the Dataset Size.}
Next, we evaluate the efficacy of our method across datasets of different sizes. We use Day to Night translation dataset, which comprises 36,728 Day images and 27,971 Night images. To understand the impact of dataset size on performance, we trained three additional models on progressively reduced subsets of the original dataset: 1,000 images, 100 images, and finally, 10 images. Table~\ref{tab:sup_num_training_images} shows that reducing the number of training images results in a slight increase in FID, but the structure preservation is largely unchanged across all different settings.  This suggests that our model can be trained on small datasets.

\begin{figure}[t!]
    \centering
    \includegraphics[width=\linewidth]{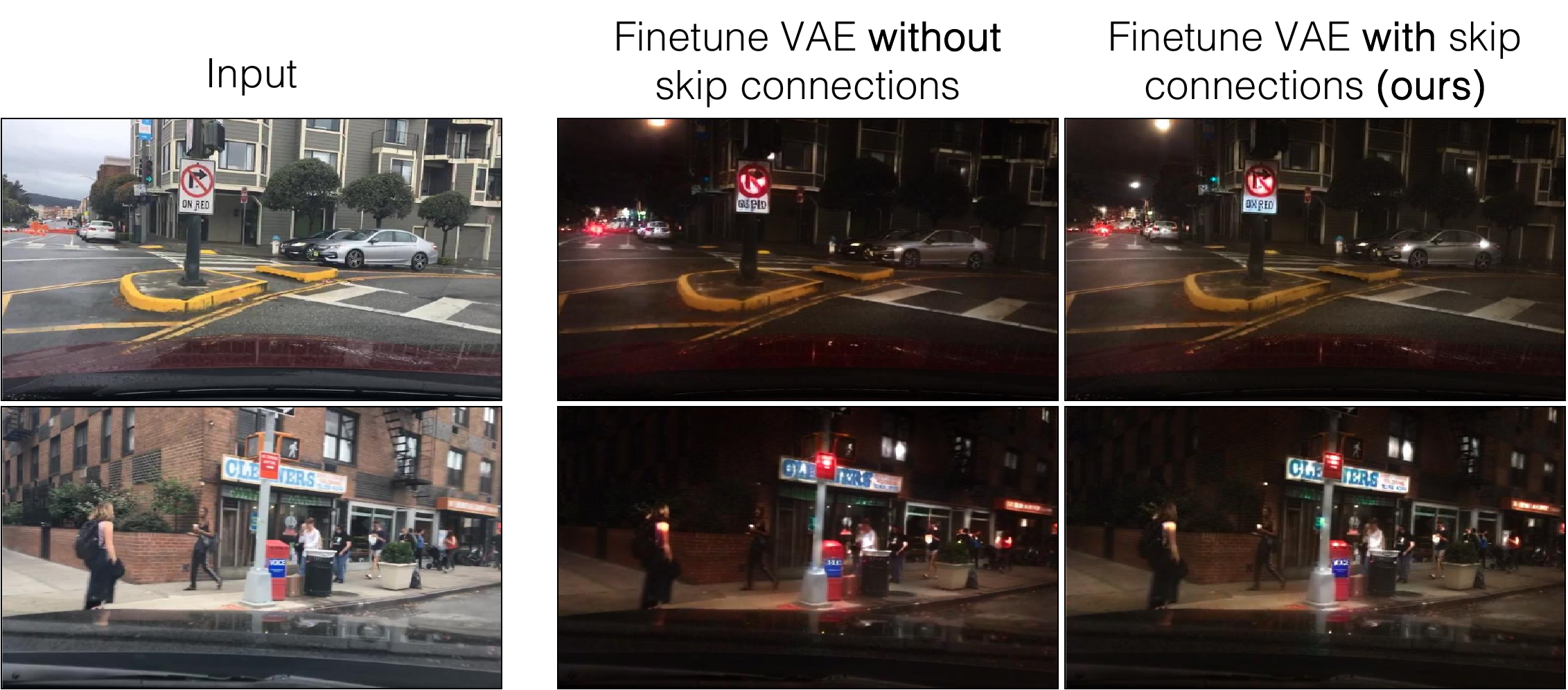}
    \caption{{\bf Finetuning encoder-decoder without skip connections}. Here we finetune the Encoder and Decoder of the VAE \textit{without} adding skip connections (middle column). Without skip connections, the method struggles to retain important details such as the text ``ON RED'' on the street sign in the top row image, the text on the store sign, and the pedestrian crossing sign in the bottom row image. In contrast, our method, with skip connections, better preserves these details. }
    \label{fig:sup_finetune_vae_noskip}
\end{figure}

\myparagraph{Role of Skip Connections.}
We additionally evaluate the role of skip connections by considering a baseline that finetunes the VAE Encoder and Decoder without adding skip connections. Figure~\ref{fig:sup_finetune_vae_noskip} shows that this baseline fails to preserve fine details such as text and street signs.

\section{Training Details} \label{sup_sec:training_details}

\myparagraph{Unpaired translation.}
For all unpaired translation evaluations, we use the four datasets listed below. For Day and Night datasets, we use 500 images from the corresponding validation at test time. The validation set for Foggy images comprises 50 images from the DENSE dataset. 
\begin{itemize}
    \item \textit{Horse $\leftrightarrow$ Zebra}: Following CycleGAN~\cite{zhu2017unpaired}, we use the 939 images form \textit{wild horse} class and 1177 images from the \textit{zebra} class in Imagenet~\cite{imagenet_cvpr09}. 
    \item \textit{Yosemite Winter $\leftrightarrow$ Summer}: We use 854 winter and 1273 summer photos of Yosemite collected from Flickr in CycleGAN~\cite{zhu2017unpaired}.
    \item \textit{Day $\leftrightarrow$ Night}: We use the Day and Night subsets of the BDD100k dataset~\cite{yu2020bdd100k} for this task.   
    \item \textit{Clear $\leftrightarrow$ Foggy:} 
    We use daytime clear images from BDD100k (12,454 images) and 572 foggy images from the `dense-fog' split of the DENSE dataset~\cite{Bijelic_2020_STF}. 
\end{itemize}
For all unpaired translation experiments, we use the Adam solver~\cite{kingma2014adam} with a learning rate of 1e-6 with a batch size of 8, $\lambda_\text{idt}=1$ and $\lambda_\text{GAN}=0.5$

\myparagraph{Paired translation.}
The training objective for the paired translation consists of three losses as mentioned in \refsec{extension} of the main paper: reconstruction loss  $\mathcal{L}_\text{rec}$ (L2 and LPIPS), GAN loss $\mathcal{L}_\text{GAN}$, and CLIP text-image alignment loss $\mathcal{L}_\text{CLIP}$. The full learning objective is shown below. We use $\lambda_\text{GAN}=0.4$, $\lambda_\text{CLIP}=4$. 
\begin{equation}\lbleq{full_paired_objective}\begin{aligned}
\arg \min_\G \mathcal{L}_{\text{rec}} + \lambda_\text{clip}\mathcal{L}_{\text{CLIP}} + \lambda_\text{GAN} \mathcal{L}_\text{GAN}.
\end{aligned}\end{equation}

We train our paired method \pairedname for two tasks: Edge2Image and Sketch2Image. Both tasks use the same community-collected dataset of artistic images~\cite{edgedataset} and follow the pre-processing of ControlNet~\cite{zhang2023adding}. 
\begin{itemize}
    \item \textit{Edge2Image.} We use a Canny edge detector~\cite{canny1986computational} with random threshold at training time. We train with Adam optimizer with a learning rate of 1e-5 for 7,500 steps with a batch size of 40. 

    \item \textit{Sketch2Image.} We generate synthetic sketches by first using a HED detector and applying data augmentations such as random thresholds, non-maximal suppression, and random morphological transformations. Our Sketch2Image is initialized with the Edge2Image model and fine-tuned for 5,000 steps with the same learning rate, batch size, and optimizer. 
\end{itemize}

\end{document}